\documentclass{article}

\usepackage{microtype}

\usepackage{graphicx}
\usepackage{subfigure}
\usepackage{booktabs} % for professional tables
\usepackage{xurl}
\usepackage{gensymb}
\usepackage{svg}
\usepackage{caption}
\usepackage{subcaption}
\usepackage{wrapfig}
\usepackage{rotating}
\usepackage{hyperref}
\usepackage{mathtools}

\usepackage{dsfont}
\usepackage{booktabs}
\usepackage{multirow}
\usepackage{numprint}

\usepackage{cellspace, tabularx}
\addparagraphcolumntypes{X}

\usepackage[accepted]{icml2024}

\usepackage{amsmath}
\usepackage{amssymb}
\usepackage{mathtools}
\usepackage{amsthm}
\usepackage{subcaption}
\usepackage[capitalize,noabbrev]{cleveref}
\usepackage{pdflscape}
\usepackage{tikz}
\usetikzlibrary{shapes, shapes.geometric, shapes.arrows, arrows, patterns, positioning, calc, arrows.meta, bending}
\usetikzlibrary{mindmap,trees}
\definecolor{mygreen}{rgb}{0.553,0.682,0.063}
\definecolor{myblue}{rgb}{0.0,0.208,0.376}
\definecolor{mygray}{rgb}{0.906,0.906,0.906}

\definecolor{darkblue}{rgb}{0.0,0.0,0.8}
\definecolor{mydarkgreen}{rgb}{0.0,0.5,0.0}
\definecolor{tablehighlight}{rgb}{0.0,0.5,0.0}

\newcommand\Reals{\mathbb{R}}
\newcommand\Xspace{\mathcal{X}}
\newcommand\Yspace{\mathcal{Y}}
\newcommand\iheight{h}
\newcommand\iwidth{w}
\newcommand\ichannels{n_c}
\newcommand\model{f}
\newcommand\tracksSet{T}
\newcommand\track{t}
\newcommand\inputInstance{I}
\newcommand\numTracksInstance{n}
\newcommand\trackMeasurement{m}
\newcommand\nTrack{{n_\track}}
\newcommand\trackMeasurementX{m_x}
\newcommand\trackMeasurementY{m_y}
\newcommand\trackMeasurementL{m_l}

\newcommand\loss{\mathcal{L}}

\newcommand*{\mycolorbox}[1]{%
\tikzstyle{mybox} = [draw=black, rectangle, inner sep=1pt, inner ysep=2pt, inner xsep=2pt, fill=white]
\tikzstyle{fancytitle} = [fill=white, text=black]
\begin{tikzpicture}
\node [mybox] (box){%
     #1
};
\end{tikzpicture}%
}
\theoremstyle{plain}

\theoremstyle{definition}

\theoremstyle{remark}

\usepackage[disable,textsize=tiny]{todonotes}
\icmltitlerunning{Estimating Canopy Height at Scale}

\begin{document}

\twocolumn[

\icmltitle{Estimating Canopy Height at Scale}



\icmlsetsymbol{equal}{*}

\begin{icmlauthorlist}
\icmlauthor{Jan Pauls}{muenster}
\icmlauthor{Max Zimmer}{zuse}
\icmlauthor{Una M. Kelly}{muenster}
\icmlauthor{Martin Schwartz}{lsce}
\icmlauthor{Sassan Saatchi}{jpl}
\icmlauthor{Philippe Ciais}{lsce}
\icmlauthor{Sebastian Pokutta}{zuse}
\icmlauthor{Martin Brandt}{copenhagen}
\icmlauthor{Fabian Gieseke}{muenster,copenhagenDIKU}
\end{icmlauthorlist}

\icmlaffiliation{zuse}{Department for AI in Society, Science, and Technology, Zuse Institute Berlin, Germany}
\icmlaffiliation{muenster}{Department of Information Systems, University of Münster, Germany}
\icmlaffiliation{lsce}{Laboratoire des Sciences du Climat et de l'Environnement, LSCE/IPSL, France}
\icmlaffiliation{copenhagen}{Department of Geosciences and Natural Resource Management, University of Copenhagen, Denmark}
\icmlaffiliation{copenhagenDIKU}{Department of Computer Science, University of Copenhagen, Denmark}
\icmlaffiliation{jpl}{Jet Propulsion Laboratory (JPL), California Institute of Technology, USA}

\icmlcorrespondingauthor{Jan Pauls}{jan.pauls@uni-muenster.de}

\icmlkeywords{Machine Learning, ICML}

\vskip 0.3in
]


% The \icmlEqualContribution command is standard text for equal contribution.w

\printAffiliationsAndNotice{}  % leave blank if no need to mention equal contribution

\begin{abstract}
We propose a framework for global-scale canopy height estimation based on satellite data.
Our model leverages advanced data preprocessing techniques, resorts to a novel loss function designed to counter geolocation inaccuracies inherent in the ground-truth height measurements, and employs data from the Shuttle Radar Topography Mission to effectively filter out erroneous labels in mountainous regions, enhancing the reliability of our predictions in those areas.
A comparison between predictions and ground-truth labels yields an MAE / RMSE of 2.43 / 4.73 (meters) overall and 4.45 / 6.72 (meters) for trees taller than five meters, which depicts a substantial improvement compared to existing global-scale maps. The resulting height map as well as the underlying framework will facilitate and enhance ecological analyses at a global scale, including, but not limited to, large-scale forest and biomass monitoring.

\end{abstract}

\section{Introduction}
\label{introduction}
Managing and conserving forest ecosystems worldwide is an indispensable component of climate adaptation and climate change mitigation strategies. Precise and up-to-date information about the health and the carbon balance of forests are, hence, critical to assess the current state of forests, to trigger appropriate countermeasures against forest loss, and to develop improved management strategies. In light of growing carbon emissions and the necessity to comply with the Paris Agreement on climate, understanding all factors affecting our climate is crucial. This includes an accurate quantification of carbon sinks to monitor carbon distribution, gains, and losses. Despite decades of interest, this quantification is still inadequate for detailed policymaking.

Traditional methods such as National Forest Inventory~(NFI) measurements, which are based on tracking and measuring individual trees, have been fundamental in estimating forest growth and loss, but are limited by their costly nature and lack of global reach. This problem is worsened by considerably varying forest monitoring efforts/techniques between nations with different financial resources~\citep{sloanForestResourcesAssessment2015}.
Advances in both Earth observation and machine learning have paved the way for the automation of forest monitoring using satellite data, including optical, radar, and LiDAR measurements, and nowadays enable more comprehensive global forest assessments~\cite{huMappingGlobalMangrove2020}.

A common way to assess the state of forests is to measure or to estimate their height, resulting in so-called canopy height maps. Such height estimates are then used to approximate the biomass and, thus, carbon, stored in the trees.
For this reason, accurate
high-resolution canopy height maps provide insights into forest cycles and dynamics and are vital for forest management and climate change mitigation.
Recent research has addressed canopy height prediction with classical machine learning approaches~\citep{potapovMappingGlobalForest2021,kacicForestStructureCharacterization2023} and deep neural networks~\citep{schwartzHighresolutionCanopyHeight2022,fayadVisionTransformersNew2023,langHighresolutionCanopyHeight2023}.
While both global and regional canopy height maps are available, there is a notable disparity in their quality.
\begin{figure}
    \centering
    \mycolorbox{\includegraphics[width=0.98\columnwidth]{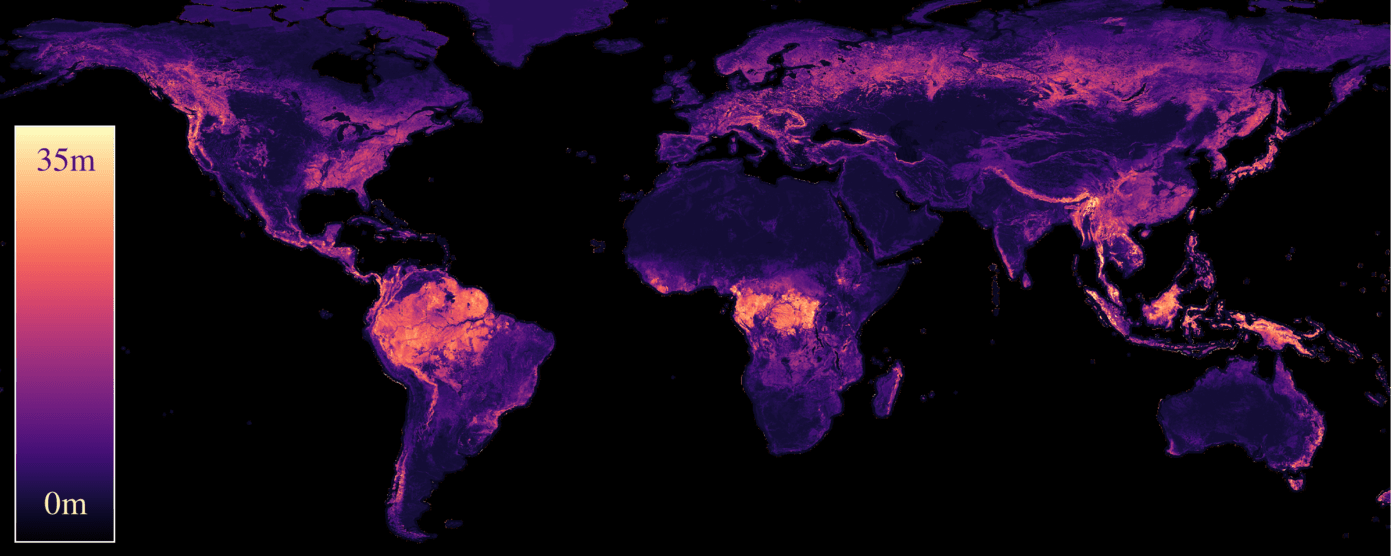}}
    \caption{A global canopy height map at a $10$~m resolution.
    }
    \label{fig:global_map}
    \vspace{-0.3cm}
\end{figure}

We address this by providing a framework for generating global-scale height maps based on satellite imagery. The overall pipeline is based on a (fully) convolutional neural network that is trained using a novel loss function to counter the label noise present in the ground-truth measurements. The framework is also based on several simple yet crucial preprocessing steps. 
We showcase the effectiveness of our approach by providing a detailed global-scale canopy height map with a resolution of $10$~m, 
see Figure~\ref{fig:global_map}.

\section{Background}
\label{sec:background}

Current canopy height estimation approaches often resort to satellite data of the \mbox{Sentinel-1}, \mbox{Sentinel-2}~\cite{ESA_Sentinel1,ESA_Sentinel2}, or the 
mission~\cite{Landsat} as input. As ground truth data, height measurements provided by the Global Ecosystem Dynamics Investigation~(GEDI) mission~\cite{dubayahGlobalEcosystemDynamics2020} are commonly used.

\subsection{Satellite Imagery}

Satellite imagery for canopy height prediction primarily comes from three missions: Landsat, Sentinel-1, and Sentinel-2. The Landsat~\cite{Landsat} program is based on a series of Earth observation satellites---operated by the NASA since 1972---and provides optical multi-spectral image data with the latest version yielding images with a pixel-resolution of $30$~m. The current revisit time is $16$ days, i.e., for a region on Earth, new data are collected every $16$ days. The satellites belonging to the Sentinel-1~\cite{ESA_Sentinel1} and the Sentinel-2~\cite{ESA_Sentinel2} missions---operated by the European Space Agency~(ESA) since 2014---offer different capabilities. The Sentinel-1 satellites feature a synthetic-aperture radar sensor, while the satellites of the Sentinel-2 mission are equipped with a multi-spectral sensor. Both types of satellites yield  imagery at a resolution of $10$~m and capture images every $6$ days at a global scale.

The aforementioned satellite missions provide new image data on a regular basis and at a global scale. Single satellite images, however, are typically not directly used for canopy height prediction due to various challenges. Radar satellites, such as the ones of the Sentinel-1 mission, can be significantly affected by heavy rain and noise in their backscatter. On the other hand, multi-spectral sensors, such as the ones of the Sentinel-2 program, often struggle with cloud and cirrus penetration. To overcome these limitations, temporal composites are usually generated, which are based on aggregating images over a specific time frame and on calculating the per-pixel median (see, e.g., Figure~\ref{fig:pipeline_overview}; top left). Such composites usually help to mitigate issues related to weather and atmospheric conditions, thus ensuring more reliable and consistent data for canopy height analyses.\footnote{
It is worth noting that airborne data are often provided at a much higher resolution (e.g., up to resolution of $10$~cm). However, these data are generally only available at a country-level, often with limited public access. This is the reason why the satellite images generally form the basis for global-scale height maps.
}

\subsection{Sparse Height Measurements}
\begin{figure}
    \centering
    \mycolorbox{\resizebox{0.95\columnwidth}{!}{\includegraphics{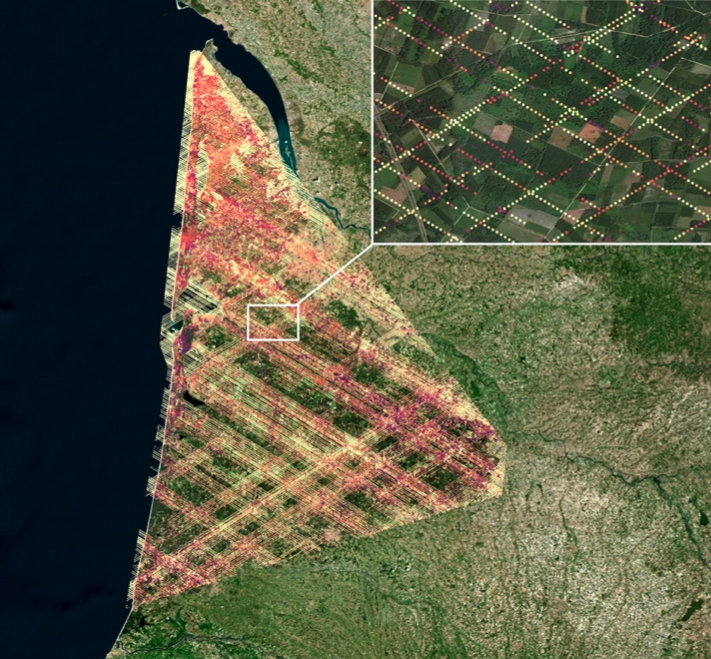}}}
    \vspace{-0.1cm}
\caption{An RGB image (based on multi-spectral image data collected by one of the Sentinel~2 satellites) along with GEDI height measurements (red/yellow dots) are shown for an area in France. 
}
\label{fig:sentinel_gedi}
\vspace{-0.3cm}
\end{figure}

The GEDI mission~\cite{dubayahGlobalEcosystemDynamics2020}, operated by the NASA, is based on a light detection and ranging~(LiDAR) module, which is installed on the International Space Station~(ISS) and which is designed to measure global vegetation height.
The system features three lasers: two ``power beams'' scanning two tracks each and one ``coverage beam'' scanning four tracks. These lasers produce measurements with a $25$~m footprint diameter, spaced $60$~m apart. The geolocation of each measurement is estimated by combining GPS and star tracker data for ISS positioning. The data acquired via these sensors essentially yield (above-ground) height measurements, i.e., for each location, the signals returned in a diameter of about $25$~m can be combined to obtain a single height value/estimate. Figure~\ref{fig:sentinel_gedi} shows a satellite image along with such (processed) GEDI height measurements. Note that 
ground-truth GEDI height measurements are only available for a fraction of the pixels.\footnote{
Over its intended two-year lifespan, GEDI was expected to scan $4$\% of the Earth's (vegetation) height, making it the most relied-upon source for canopy height measurements due to its global coverage.}

\subsection{Generating Height Maps}
\begin{figure}[t]
    \centering
    \includegraphics[width=\columnwidth]{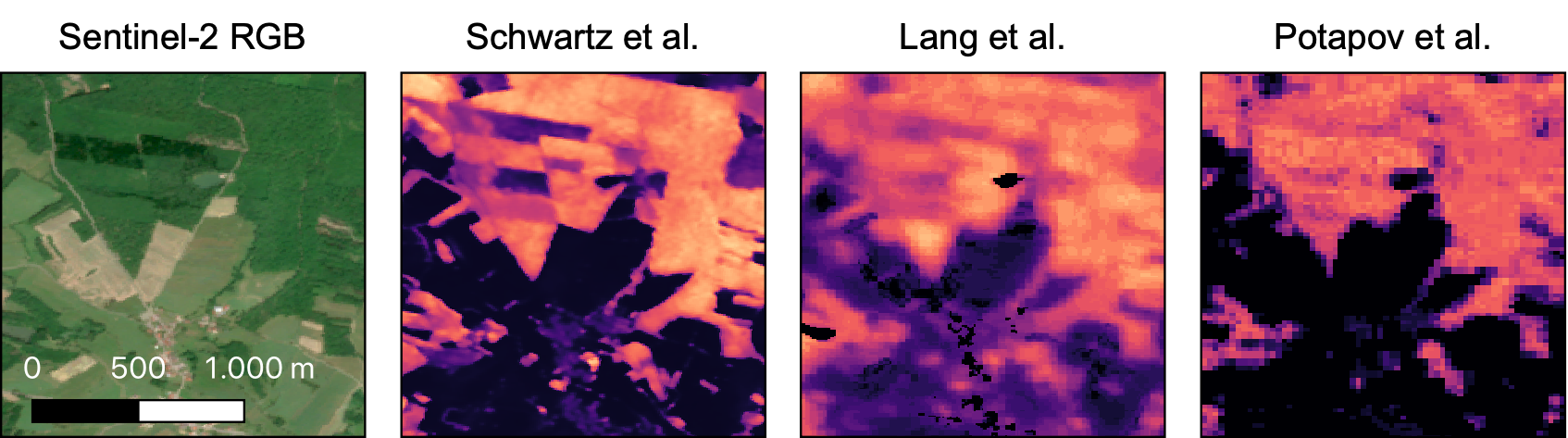}
    \vspace{-10pt}
    \caption{Visual comparison of a regional map \citep{schwartzFORMSForestMultiple2023} and two global maps \citep{langHighresolutionCanopyHeight2023,potapovMappingGlobalForest2021} (heights from $0$~m to $35$~m; see \cref{fig:global_map} for the colormap).}
    \label{fig:regional_difference}
    \vspace{-0.3cm}
\end{figure}
The goal of a height estimation model is to provide a height estimate for each location/pixel on Earth/in a given satellite image, see again Figure~\ref{fig:sentinel_gedi}. The output of three height models is shown in Figure~\ref{fig:regional_difference}. In all three cases, satellite imagery was used as input data.
\citet{potapovMappingGlobalForest2021} proposed the first high-resolution ($30$~m) global canopy height map in 2011 using satellite imagery of the Landsat program~\cite{Landsat}. To obtain a height estimate per pixel, a random forest model was trained using GEDI height measurements.
More recently, fully convolutional neural networks and vision transformers have been used to generate height maps at a country-level using Sentinel-1/-2 data, such as the maps provided by \citet{schwartzFORMSForestMultiple2023} and by \citet{fayadVisionTransformersNew2023}, which both exhibit a resolution of $10$~m.\footnote{Such data products are often based on manual preprocessing steps that are tailored to the specific needs of the region at hand.} Using similar techniques, \citet{liuOverlookedContributionTrees2023} developed a canopy height map for Europe using 3m resolution commercial satellite imagery and so-called airborne laser scanning~(ALS) data as ground truth. The most recent wall-to-wall (i.e. with global coverage) height map was provided by \citet{langHighresolutionCanopyHeight2023}, who also used Sentinel-1/-2 data with a resolution of $10$~m and GEDI labels for training.

Hence, several regional~\cite{schwartzFORMSForestMultiple2023,fayadVisionTransformersNew2023,liuOverlookedContributionTrees2023} as well as two wall-to-wall height maps~\cite{potapovMappingGlobalForest2021,langHighresolutionCanopyHeight2023} have been proposed in the recent past. Note that the quality of regional maps is generally higher compared to one of the global maps. In this work, we close this quality gap by providing a wall-to-wall canopy height map with a resolution of 10~m, whose quality is comparable with the ones of the local products, while providing height estimates at a global scale.

\section{Approach}
\label{methodology}
\begin{figure*}[t]
    \centering
    \includegraphics[width=\textwidth]{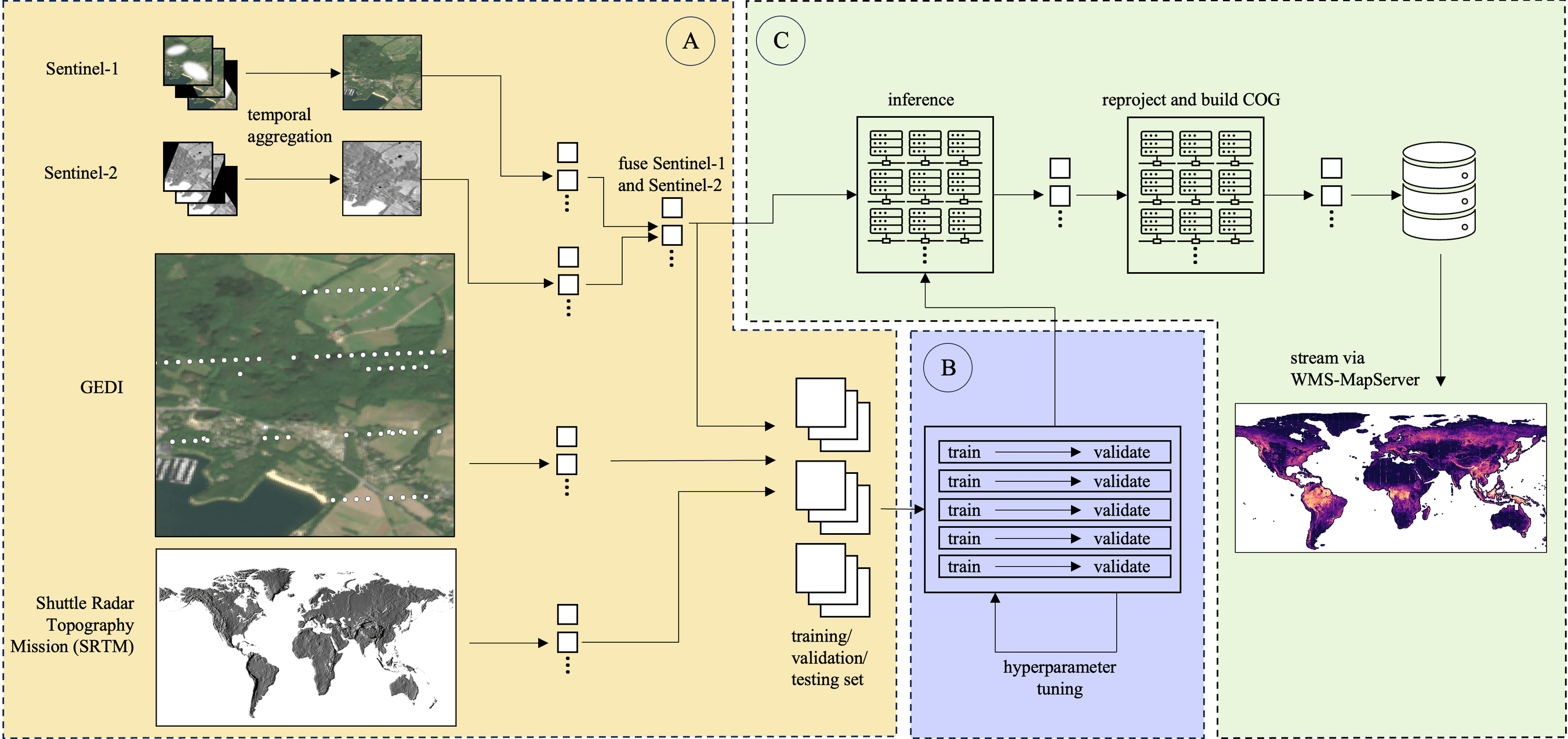}
    \caption{Our approach to generate height maps at a global scale: (A) data collection and preprocessing, (B) model training, and (C)~global-scale inference. Initially, image composites based on Sentinel-1 and Sentinel-2 imagery are constructed and corresponding GEDI measurements as well as SRTM data are collected.  
    After training the model using hyperparameter grid search, it is applied in a distributed manner, with the canopy height estimates being subsequently reprojected and made available through streaming services. 
    }
    \label{fig:pipeline_overview}
\end{figure*}

We introduce a methodology designed to produce high-quality, high-resolution global forest canopy height maps, addressing the previously mentioned challenges. Selecting appropriate input and label data is crucial. Next, we detail the corresponding selection and preprocessing process, our model architecture, and the training approach. The overall pipeline is provided in Figure~\ref{fig:pipeline_overview} and comprises the following steps: (A) data collection and preprocessing, (B) model training, and (C) global-scale inference.

\subsection{Data Collection and Preprocessing (A)}\label{section:data}
For the Sentinel-1 data, we adopt a methodology similar to that used by \citet{schwartzHighresolutionCanopyHeight2022}, i.e., we calculate the per-pixel median for images taken during the summer leaf-on season (April to October 2020 in the northern hemisphere and October 2019 to April 2020 in the southern hemisphere). For the Sentinel-2 data, a distinct method is necessary due to significant cloud coverage in some areas, especially in rainforest regions. Here, we resort to a cloud reduction algorithm adapted from \citet{justinbraatenSentinel2CloudMasking} to minimize cloud artifacts, cloud shadows, and cirrus in the data. In total, we consider $14$ channels/bands ($4$ from Sentinel-1 and $10$ from Sentinel-2). We refer the reader to \cref{cloudmasking} for the implementation details and to \cref{examples_of_images} for examples.
As ground-truth labels, we resort to the data provided by the GEDI mission. In particular, we make use of the so-called RH100 metric, which measures the height at which $100$\% of the returned signals are registered. To enhance the quality of these labels, we consider various filtering steps, such as distinguishing between night and daytime shots to avoid solar radiation disturbing the height measurements. The total volumes of the preprocessed Sentinel-1 and Sentinel-2 images and the GEDI data are $45$~TB and $1$~TB, respectively.

\begin{figure}[t]
    \centering
    \includegraphics[width=0.92\columnwidth]{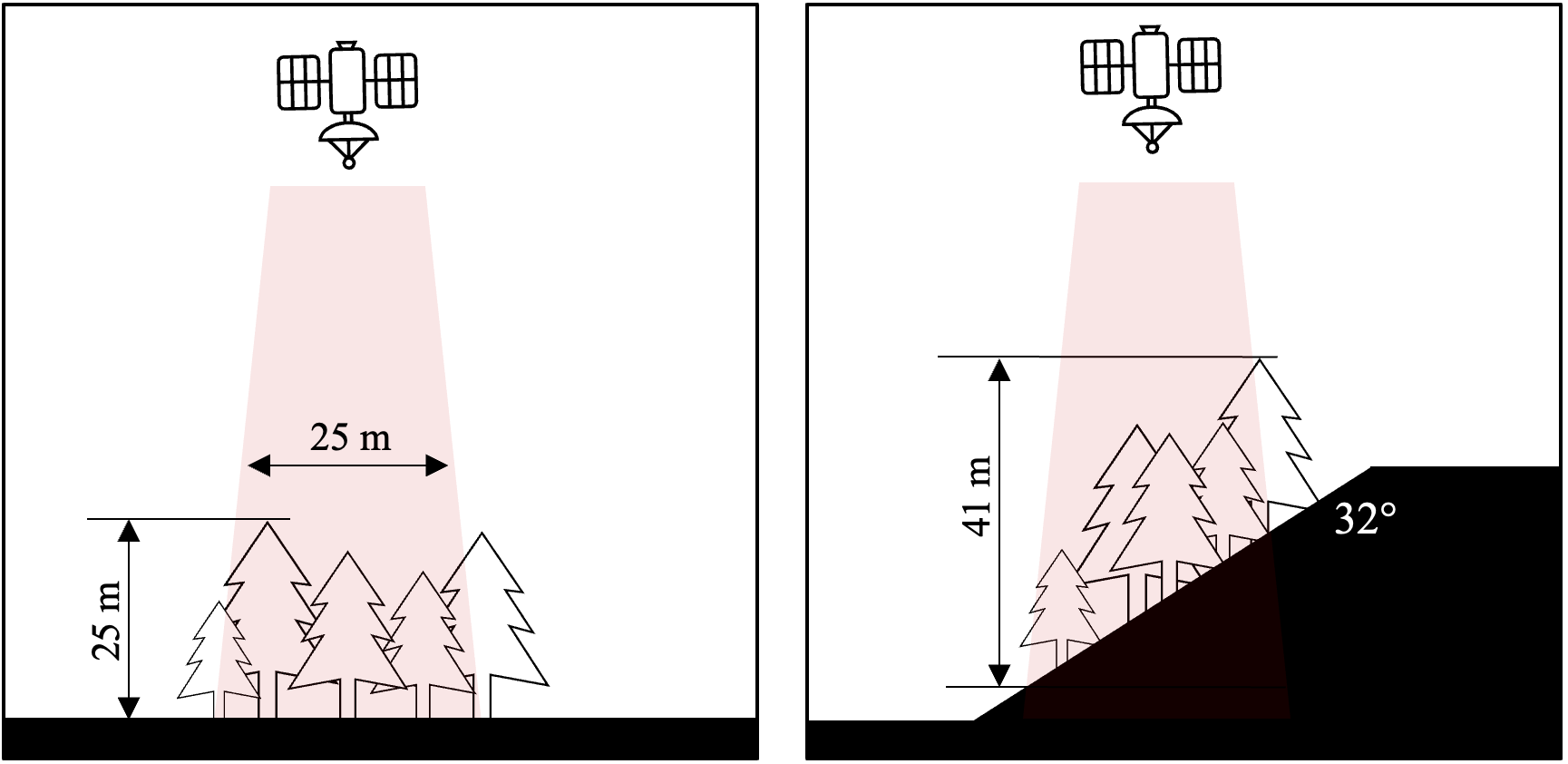}
    \vspace{-3pt}
    \caption{
    The GEDI instrument records the signals returned from the ground within a $25$~m diameter. The height is then essentially computed based on the differences of the first and last signals. While satisfying height values are obtained for most areas, this often leads to inaccurate height measurements for slopes.
    }
    \label{fig:gedi_slope}
    \vspace{-0.3cm}
\end{figure}
Typically, in the context of GEDI, the canopy height is calculated as the difference between the first and last contact point of the signal (see \cref{fig:gedi_slope}). In areas with steep slopes, however, this approach often inaccurately treats unforested slopes as high canopy or leads to a significant overestimation of tree height, see \cref{fig:gedi_slope} for an illustration of the latter issue. To mitigate the impact induced by incorrect canopy height values in mountainous areas, we resort to data from the Shuttle Radar Topography Mission~(SRTM), which has mapped the Earth's surface topography with approximately $1$~arc second resolution (about $30$~m). It is important to note the distinction between the surface return, which is the initial contact point for measurements (such as tree canopies or building surfaces), and the terrain return, which is the actual ground level. The SRTM data, capturing the surface return, show that forest edges reflect a height change equal to the tree height, thereby complicating slope measurement filtering. Hence, for training, we exclude all GEDI measurements where the corresponding SRTM slope exceeds $20$\degree.\footnote{As we do not want to discard any valid measurement, we apply a $20$\degree filter threshold (a forest edges with trees up to $40$~m in height already corresponds to a slope of $15$\degree).}

Overall, we generate data samples by selecting satellite image tiles uniformly at random. 
For each image, we extract (the same number of) patches of size $512 \times 512$ pixels and for each such patch,
we extract the associated GEDI height measurements as labels, resulting in about $10$ to $400$ height measurements per patch. Overall, $\numprint{100000}$ patches are extracted, out of which $80$\% are used for training, $10$\% for validation, and $10$\% for testing. 
Patches extracted from an image tile belong to only one of these three sets, ensuring no overlap.
More details are provided in the appendix.

\subsection{Model Training (B)}
\label{model_architecture_and_training}

Next, we describe the model architecture and the technical details of the training and hyperparameter tuning process.

\subsubsection{Model Architecture \& Parameters}
Let $\Xspace=\Reals^{\ichannels\times\iwidth\times\iheight}$ be the input space containing images with $\ichannels$ channels (in our case $\iwidth=512$, $\iheight=512$, and $\ichannels=14$) and let $\Yspace=\mathbb{R}^{\iwidth\times\iheight}$ be the output space. The goal is to train a height estimation model $\model: \Xspace \rightarrow \Yspace$ that assigns one height estimate to each of the input pixels.
Given the global application scope, achieving a balance between prediction accuracy and inference time is essential.

Following \citet{schwartzHighresolutionCanopyHeight2022}, we resort to the well-known U-Net architecture~\citep{ronnebergerUNetConvolutionalNetworks2015}, which is a semantic segmentation model. 
We set the number of output classes to one and make use of a linear activation for the final layer in order to estimate the GEDI heights. 
We also replace the original architecture's backbone by a ResNet50~\citep{heDeepResidualLearning2015} backbone. 
We optimize the model weights using the AdamW~\citep{AdamW} optimizer with a weight decay of $0.001$, a batch size of $32$, and an initial learning rate of $0.001$. We also resort to a linear learning rate warm-up for the first $10$\% of the iterations and a linear learning rate scheduler for the remaining $90$\% of the iterations. To address the skewed label distribution (also see Section~\ref{sec:error_analysis}) in our training set, a weighted sampler is used. The validation set was used for hyperparameter selection. We refer to the appendix for more details related to the hyperparameter search and the training process.

\subsubsection{Shift-Resilient Loss}
\begin{figure}
    \centering
    \mycolorbox{\includegraphics[width=0.98\columnwidth]{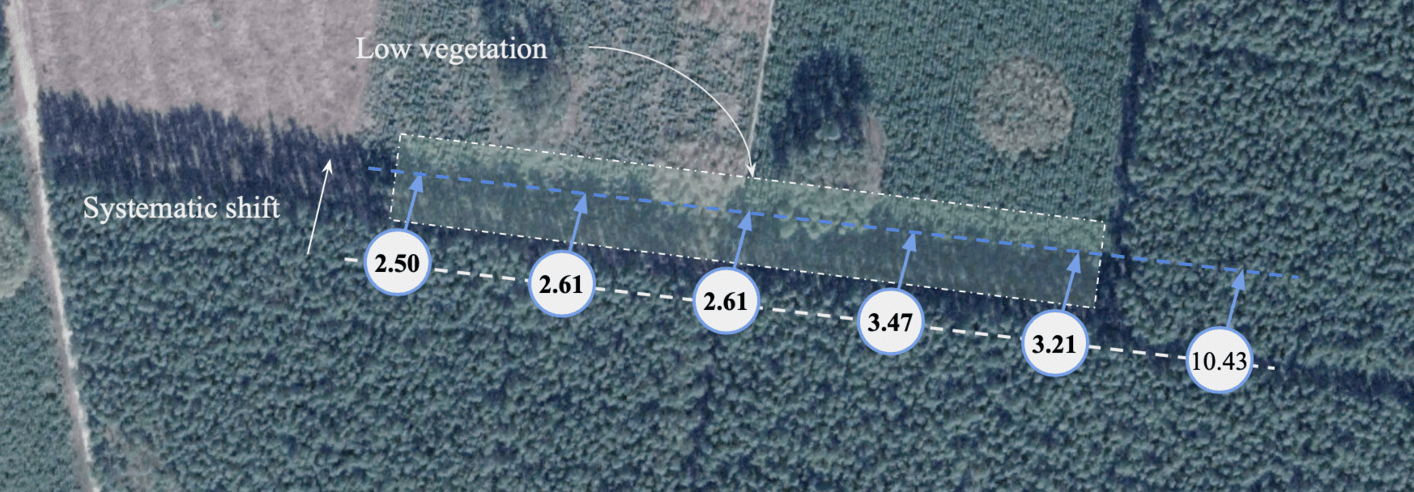}}
    \vspace{-0.15in}
    \caption{The geolocations of GEDI measurements often exhibit a systematic shift, leading to faulty ground-truth labels (each circle shows a height measurement; background image: Google Maps).}
    \label{fig:gedi_shift}
    \vspace{-8pt}
\end{figure}
When inspecting GEDI measurements, it can be observed that their geolocation is not always precise, i.e., a systematic pattern in geolocation errors is often given in the data, characterized by consistent shifts in each track of measurements, see \cref{fig:gedi_shift}. These shifts are problematic since correct predictions made by a model might be considered to be wrong by standard loss functions.
A recent approach employs a high-resolution (1m) digital terrain model~(DTM) to individually correct GEDI measurements based on their ground return~\cite{schleichImprovingGEDIFootprint2023}. While this method enhances the geolocation accuracy of GEDI measurements, it is not feasible for global-scale applications due to a lack of up-to-date DTMs at a global scale. 

We propose a shift-resilient loss function tailored to this learning scenario that can be added on top of any other regression loss function. 
The basic idea is that we allow each track to be shifted in any direction by a magnitude smaller than a radius $r>0$ and select the shift with the lowest associated loss. 
More specifically, for a given input instance $\inputInstance \in \Xspace$, let $\tracksSet_\inputInstance=\{\track_1, \ldots, \track_\numTracksInstance \}$ be the corresponding set of GEDI tracks containing the (potentially shifted) ground-truth height measurements. Here, each of the tracks $\track \in \tracksSet_\inputInstance$ is composed of a set $\track=\{\trackMeasurement_1, \ldots, \trackMeasurement_\nTrack\}$ of ground-truth measurements and each such measurement $\trackMeasurement \in \track$ is represented by $\trackMeasurement=(\trackMeasurementX, \trackMeasurementY, \trackMeasurementL)$, where $\trackMeasurementX$ and $\trackMeasurementY$ depict the pixel coordinates and $\trackMeasurementL$ the height value.
Note that the measurements belonging to one track are on a ``line''. For example, the ``GEDI'' input image patch on the left hand side of Figure~\ref{fig:pipeline_overview} contains five tracks, where each white dot on a track represents one measurement. 

For a track $\track \in \tracksSet_\inputInstance$ in a set of tracks $\tracksSet_\inputInstance$ associated with an input image~$\inputInstance \in \Xspace$, we define $\track_\inputInstance \in \Yspace$ as
\begin{equation}
    \track_\inputInstance(x,y)  \coloneq
    \begin{cases}
        \trackMeasurementL & \text{if } \exists \ \trackMeasurement \in \track \text{ s.t. } (x,y)\!=\!(\trackMeasurementX,\trackMeasurementY),\\
        0              & \text{otherwise.}
    \end{cases}
\end{equation}
For each such $\track_\inputInstance$, we also define a $\track^0_\inputInstance \in \{0,1\}^{\iwidth \times \iheight}$ as 
\begin{equation}
    \track^0_\inputInstance(x,y)  \coloneq
    \begin{cases}
        1 & \text{if } \track_\inputInstance(x,y) \neq 0,\\
        0              & \text{otherwise.}
    \end{cases}
\end{equation}

For an $\inputInstance \in \Xspace$ with corresponding height estimates $\model(\inputInstance)$, the (non-shifted) loss $\loss_{\textit{NS}}$ can be computed as
\begin{equation}
\label{eq:non_shifted_loss}
    \mathcal{L}_{\textit{NS}} \left(\model(\inputInstance), \tracksSet_\inputInstance\right)  \coloneq \frac{1}{N}\sum_{\track \in \tracksSet_\inputInstance}\loss\left(f(I)\odot \track_\inputInstance^0, t_{I}\right),
\end{equation}
where $N=\sum_{\track \in \tracksSet_\inputInstance} |\track|$,~$\odot$ the element-wise Hadamard product, and $\loss:\Yspace \times \Yspace \rightarrow \Reals^+$ a standard (pixel-wise) loss for regression (e.g., $L_2$ loss). 

As mentioned above, the geolocations of the GEDI measurements are often not precise. Hence, the loss function $\loss_{\textit{NS}}$ is not an ideal choice to assess the quality of $\model(\inputInstance)$ since $\model(\inputInstance)$ would be compared with ground-truth values inaccurate in their geolocation, potentially leading to a high loss although the model output might be correct. Note, however, that the shift is (relatively) consistent accross all the GEDI measurements belonging to the \emph{same} track.
Hence, to compensate for possible (unknown) systematic geolocation shifts $\delta = (\delta_ x, \delta_y) \in \mathbb{Z}^2$ of the GEDI measurements belonging to the same track, we define the following shifted ground-truth targets $\track_{\inputInstance, \delta}(x,y) \coloneq \track_\inputInstance(x-\delta_ x,y-\delta_ y)$ 
(measurements close to any of the borders are handled appropriately). Then, we consider the following (shifted) loss function:
\begin{eqnarray}
\label{eq:shifted_loss}
    \loss_{\textit{S}} \left(\model(\inputInstance), \tracksSet_\inputInstance\right) \!\!\!\!\! & \coloneq&\!\!\!\!\!\frac{1}{N}\sum_{\track \in \tracksSet_\inputInstance}\min_{\delta}\mathcal{L}\left(\model(\inputInstance)\odot t_{\inputInstance,\delta}^0, \track_{\inputInstance,\delta}\right)  \\
    &\textrm{s.t.}& \sqrt{\delta_ x^2 + \delta_ y^2} \leq r . \nonumber
\end{eqnarray}
Thus, all shifts meeting the constraint are considered, and the one with the lowest pixel-wise loss is chosen.\footnote{In case there are fewer than ten measurements available in a track $t_I$, we do not allow shifting for that track, as it is hard to estimate the systematic shift with only a few measurements. 
}
In our case, we choose the pixel-wise Huber loss for $\loss$, as it is more robust to outliers than the $L_2$ loss. The variable $r$ is a user-defined parameter specifying the allowed shift radius. For the GEDI measurements, it makes sense to restrict $r$ to about $\sqrt{2}$ as $80.8\%$ of the GEDI tracks exhibit a mean geolocation error below 10 meters~\cite{tangEvaluatingMitigatingImpact2023}.

\subsection{Global-Scale Inference (C)}

After training, the model is deployed globally by partitioning the Earth into patches of size $312 \times 312$ pixels. For predictions, we consider image patches of size $512 \times 512$ pixels, i.e., we include \numprint{100} pixels at each border for context (the predictions within these borders are ignored for the final map/mosaic). 
The computational process consumes approximately $\numprint{1500}$ GPU hours on a GPU cluster with various GPU devices (e.g., Nvidia RTX3090s and A100s).

For an efficient visualization of the map, we resort to several postprocessing steps. Initially, we standardize the map projection across all predictions to the EPSG:3857 (Web Mercator) projection. Subsequently, we convert predictions into the cloud optimized GeoTiff~(COG) format. This conversion step involves, among other things, data compression and the creation of high-level overviews. These steps are computationally very demanding, with reprojection and COG creation requiring about $\numprint{81000}$ CPU hours.

Finally, we resort to a web map service~(WMS) to stream the canopy height map to any geographic information system software. We visualize the map using a magma color ramp, a common choice for depicting canopy heights. A high-level overview of the final map is presented in \cref{fig:global_map} (a more detailed version can be found in the appendix).

\subsection{Source Code \& Canopy Height Map}
Our source code, as well as detailed documentation, are publicly available on GitHub.\footnote{\href{https://github.com/AI4Forest/Global-Canopy-Height-Map}{https://github.com/AI4Forest/Global-Canopy-Height-Map}} The induced global canopy height map is accessible through the Google Earth Engine (see Appendix~\ref{sec:global_canopy_height_map}).

\section{Results}

\begin{table}
    \caption{Comparison of our height map with two publicly available global-scale height maps~\cite{potapovMappingGlobalForest2021,langHighresolutionCanopyHeight2023}.}
    \label{tab:metrics_global}
    \vskip 0.1in
    \centering
    \resizebox{\columnwidth}{!}{
    \begin{sc}
    \begin{tabular}{clcccc}
    \toprule
        Filter & Metric & Unit & Lang & Potapov & Ours \\ \midrule
        & MAE & $m$ & $\phantom{00}6.47$ & $\phantom{00}6.92$ &$\phantom{0}2.43$\\
        & MSE & $m^2$ & $\phantom{0}74.70$ & $\phantom{0}85.58$ & $22.41$\\
       no & RMSE & $m$ &$\phantom{00}8.62$ & $\phantom{00}9.25$ & $\phantom{0}4.73$\\ 
        & RRMSE &  & $\phantom{00}1.39$ & $\phantom{00}2.38$ & $\phantom{0}0.53$\\ 
        & MAPE &  & $\phantom{00}1.01$& $\phantom{00}0.97$ & $\phantom{0}0.22$\\\midrule
        & MAE & $m$ &$\phantom{00}8.80$ & $\phantom{0}10.01$ & $\phantom{0}4.45$\\
        & MSE & $m^2$ &$121.51$ & $154.45$ & $45.12$\\
        $\trackMeasurementL > 5$  & RMSE & $m$ & $\phantom{0}11.02$ & $\phantom{0}12.43$ & $\phantom{0}6.72$\\
        & RRMSE&  & $\phantom{00}1.02$ & $\phantom{00}1.77$ & $\phantom{0}0.49$\\
        & MAPE & & $\phantom{00}0.76$ & $\phantom{00}0.77$ & $\phantom{0}0.28$\\\bottomrule
    \end{tabular}
    \end{sc}
    }
    \vspace{-0.15in}
\end{table}

We provide a comparison of our height map with two other global height map products commonly used in the field~\cite{potapovMappingGlobalForest2021,langHighresolutionCanopyHeight2023} and with one regional height map that has recently been produced for France~\cite{schwartzFORMSForestMultiple2023}. 
We also analyze the errors still made by our model and the impact of key filtering steps.

\begin{figure*}[t!]
    \centering
    \includegraphics[width=1\textwidth]{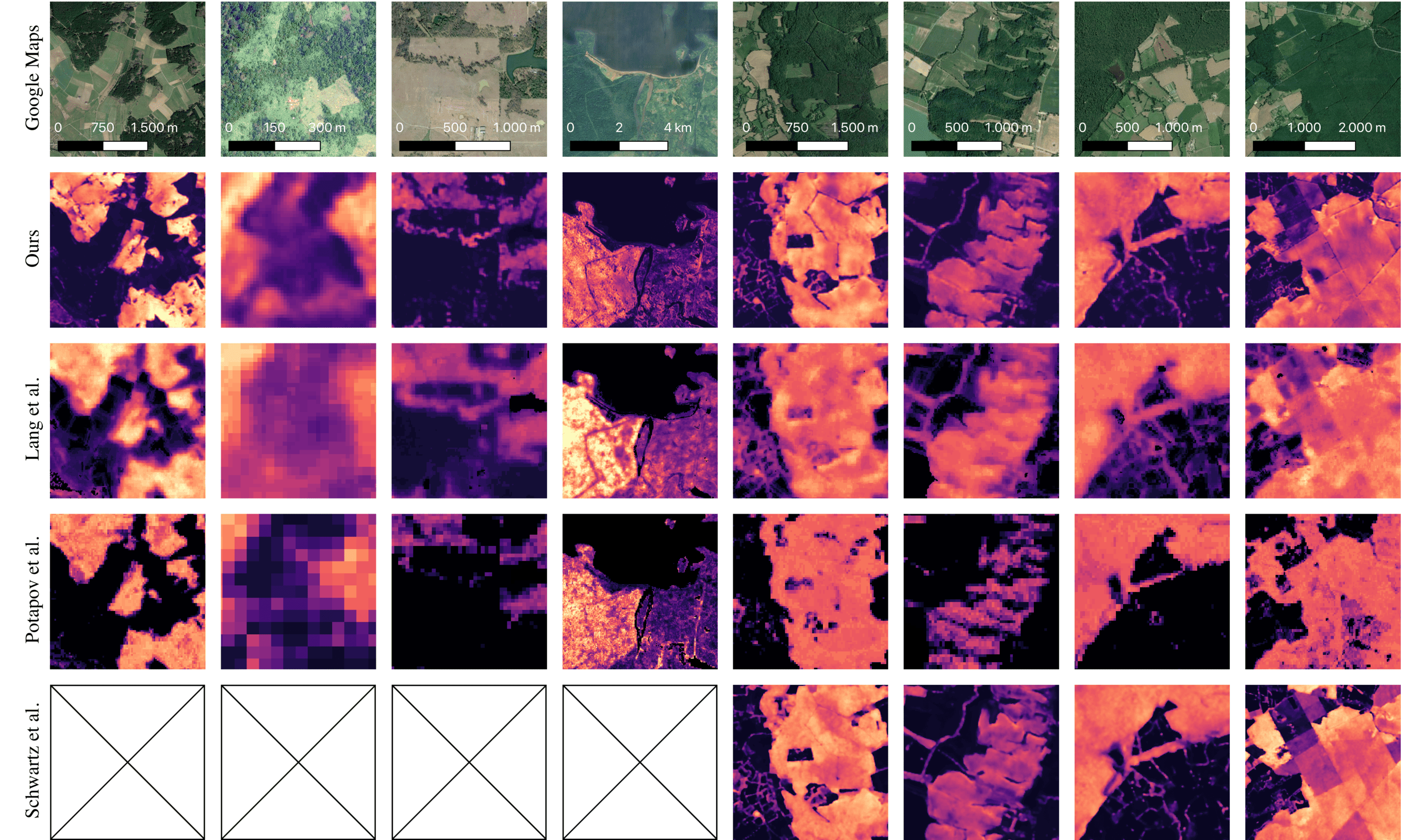}
    \caption{Visual comparison of our map with two global height maps (first four columns) \citep{langHighresolutionCanopyHeight2023,potapovMappingGlobalForest2021} and a regional map for France \citep{schwartzFORMSForestMultiple2023} (last four columns). The first row provides high-resolution Google Maps imagery for visual context. Our canopy height map shows an improvement compared to existing global canopy height maps in terms of clarity and detail for all eight examples. The visual quality of our map for the four examples in France is close to the one of the regional map (heights from $0$~m to $35$~m; see \cref{fig:global_map} for the colormap).
    }
    \label{fig:visual_comparison}
\end{figure*}
\subsection{Comparison of Height Maps}
We resort to several performance metrics as well as to visual examples in order to compare the quality of the height maps.
\subsubsection{Metric-Based Comparison}
We consider the mean absolute error (MAE), the mean squared error (MSE), the root mean squared error, the relative root mean squared error (RRMSE), and the mean absolute percentage error (MAPE) to asses the quality of the different height map products (all maps are compared using the same non-shifted GEDI data). All numbers the provided are obtained using the test area (see Appendix~\ref{appendix:train_val_test}). No test instances have been used for training and model selection.

The results are shown in \cref{tab:metrics_global}. 
As \citet{langHighresolutionCanopyHeight2023} and \citet{potapovMappingGlobalForest2021} both use a forest mask, the metrics are shown with no filters applied as well as only for GEDI measurements larger than $5$~m.
The top part of the table presents the results without any additional filter being applied to the GEDI data. It can be seen that our height map yields a significantly smaller MAE, standing at $2.43$~m, compared to the maps of \citet{langHighresolutionCanopyHeight2023} and \citet{potapovMappingGlobalForest2021}, which result in an MAE of $6.47$~m and $6.92$~m, respectively. Similar improvements are given for the other metrics. Note that these values do not account for the fact that a GEDI label of $3$~m corresponds to the ground, but most height maps resort to forest masks in order to set the corresponding height values to $0$~m. This results in a consistent $3$~m error for all accurately classified ground predictions. To address this, we also provide a comparison using only the GEDI height measurements larger than $5$~m ($\trackMeasurementL>5$). For these filtered data, a similar quality gain of our map compared to the other two existing height map products can be observed.

\subsubsection{Visual Comparison}
Next, we compare the maps using various visual examples. Here, visual quality refers to a map's appearance, particularly at forest edges, patches, and in terms of ``resolution''.
Some examples are given in \cref{fig:visual_comparison}. In addition to the two global height maps~\cite{langHighresolutionCanopyHeight2023,potapovMappingGlobalForest2021}, we also consider the regional map for France recently proposed by~\citet{schwartzFORMSForestMultiple2023}. 

We start with a comparison of the two global map products and our map. 
The map by \citet{langHighresolutionCanopyHeight2023} generally yields a smoother texture compared to our map, leading to some loss of detail. For instance, it sometimes fails to accurately depict forest structure (as it can be seen, e.g., in the first column). The map also yields the highest predictions, which sometimes leads to fewer details in low-height areas (visible in, e.g., the fourth column). 
The map of \citet{potapovMappingGlobalForest2021} is derived from Landsat data with a resolution of 30~m. Hence, naturally, it appears to be more coarse compared to the other maps (which are based on Sentinel-1/Sentinel~2 imagery with a resolution of $10$~m). It also struggles with capturing forest variances. In particular, it often renders forest areas using an almost uniform height and, thus, does not capture smaller forest patches and fine structural details. This is, for instance, visible in the fifth and the last column of the table. In contrast, our map often more accurately identifies fine structural details, such as the pathways and small forest patches (see, e.g., the first and the fifth column).

Comparing our height map with the regional map for France provided by \citet{schwartzFORMSForestMultiple2023}, we observe notable similarities between both maps. 
Compared to the other two global maps, the most significant distinction lies in the ability of these two maps to identify fine structures such as forest gaps, roads within forests, and adjacent forest areas. For instance, in the fifth column, our map successfully identifies multiple gaps and roads, a detail both global maps do no capture. Notably, even the regional map fails to accurately depict the gap line in the upper half of the image.\footnote{This enhanced detail in our map could be attributed to the shifted loss function, which makes the model more resilient against the label noise present in the ground-truth GEDI measurements. This seems to effectively reduce smoothing at ambiguous locations.}

\begin{figure}[t]
    \centering
    \includegraphics[width=1.0\columnwidth]{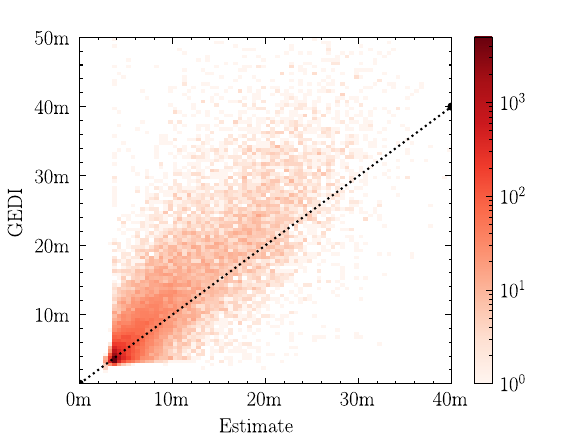}
    \vspace{-0.15in}
    \caption{Scatter plot showing height estimates versus ground-truth GEDI height measurements ($R^2$: $0.72$; log scale).}
    \label{fig:scatterplot}
    %\vspace{-0.1in}
\end{figure}

\subsection{Error Analysis \& Filtering Steps}
\label{sec:error_analysis}
Next, we analyze the remaining errors in more detail as well as the effect of the SRTM filtering step.

\subsubsection{Underestimation of High Heights}
In Figure~\ref{fig:scatterplot}, a scatter plot comparing GEDI ground-truth measurements and our height estimates is provided. The logarithmic scale of density in the scatter plot illustrates the label imbalance present in the GEDI data, with a mean height of $6.33$~m and a standard deviation of $7.17$~m. Measurements at $40$~m are approximately $423$ times rarer than those at $3$~m, and $90$\% of the measurements are below $14$~m, showing a pronounced bias towards lower height measurements.
The use of the shifted loss function significantly reduces completely incorrect predictions (i.e., errors ``located'' close to the x- or y-axis), which are typically prevalent in canopy height maps. However, our model still underestimates the ground-truth measurements, which is especially the case for high heights (e.g., $\trackMeasurementL>30$~m). Note that this is a common problem across the available height maps.

\begin{figure}
    \centering
    \includegraphics[width=\columnwidth]{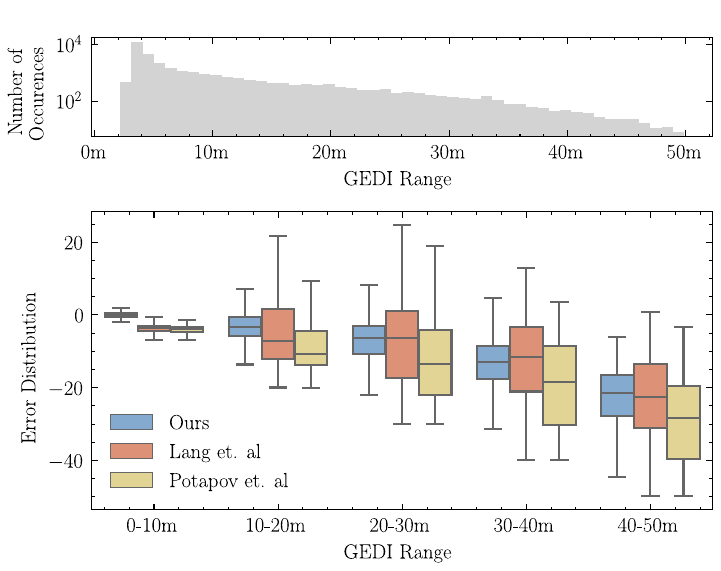}
    \caption{Top: Distribution of GEDI measurements (log scale). Bottom: Evaluation of the errors given different height ranges from $0$~m to $60$~m. Here, negative errors indicate that the estimates are smaller than the ground-truth GEDI measurements.}
    \label{fig:error_height_distribution}
\end{figure}

\cref{fig:error_height_distribution} presents boxplots for all three global maps, where the errors are sorted according to the ground-truth GEDI height measurements in bins of $10$~m.
Our map shows significantly lower error rates for vegetation heights under $20$~m compared to the other two maps. For heights up to $10$~m, the error rate for the map of \citet{potapovMappingGlobalForest2021} is comparable to that of the map of~\citet{langHighresolutionCanopyHeight2023}. However, for heights greater than $10$~m, it deteriorates, generally falling $5$--$10$~m short of both other maps. While our map shows an improved accuracy for heights up to $20$~m, this advantage does not extend to greater heights, where our errors align more closely with those of \citet{langHighresolutionCanopyHeight2023}.\footnote{The approach of \citet{langHighresolutionCanopyHeight2023} strongly aimed at addressing the underestimation of high heights/tall trees, which results in the overestimation of smaller heights, as indicated by the $+20$~m whiskers in the $10$~m to $30$~m height ranges.} Across all height bins, our map consistently shows the lowest variance, particularly notable in trees up to $30$~m, where the reduction in variance is significantly prominent.

\begin{figure}[t]
    \centering
    \includegraphics[width=\columnwidth]{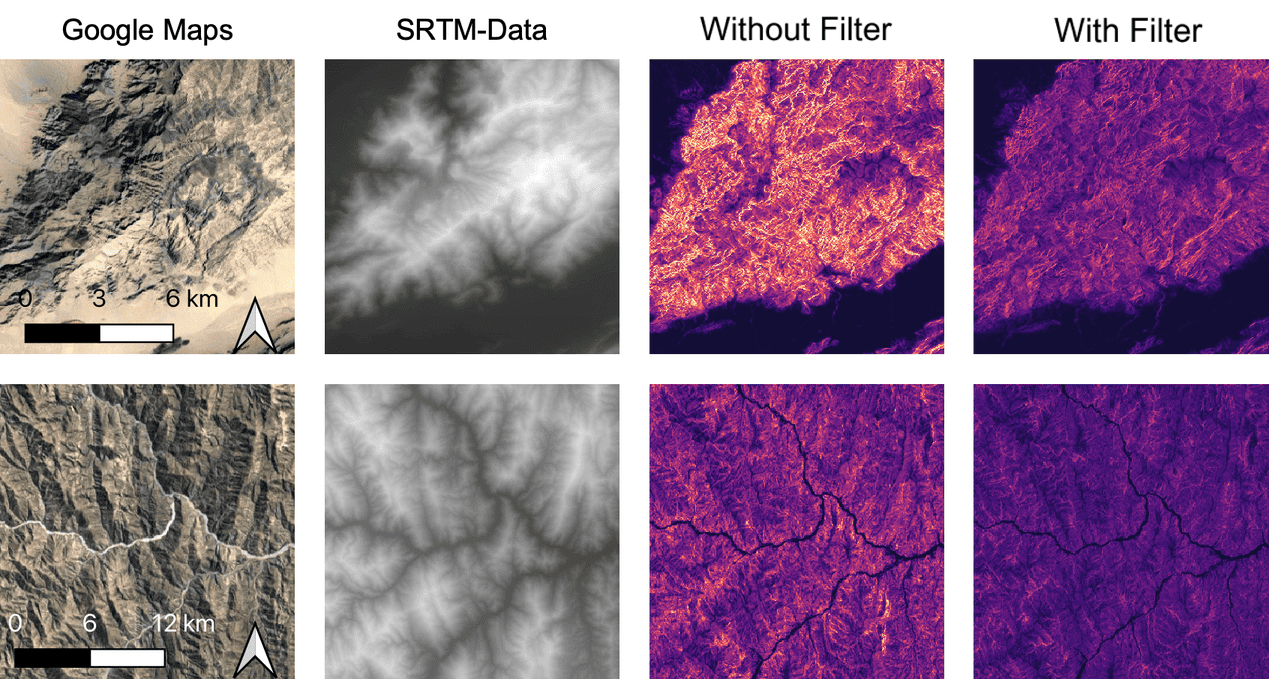}
    \caption{Visual illustration of the impact of the SRTM filtering step. The SRTM filtering drastically reduces the overestimation of high heights in mountainous areas (heights from $0$~m to $35$~m; see \cref{fig:global_map} for the colormap). 
    }
    \label{fig:visual_comparison_mountains}
\end{figure}

\subsubsection{SRTM Filtering for Mountainous Regions}
We resort to a filtering step for the GEDI height measurements for mountainous regions using SRTM data, see Section~\ref{section:data}. 
While the SRTM provides valuable surface topography data, it is important to note that it captures surface elevation rather than pure terrain elevation. Our use of SRTM data is, due to this limitation, not a perfect solution for improving height estimation in mountainous regions, see \cref{fig:visual_comparison_mountains}. However, incorporating SRTM data still represents an improvement over not addressing the topographical challenges at all. Hence, the implementation of a filter based on SRTM data predominantly ensures the exclusion of unrealistically high canopy predictions in steep terrains.

\section{Conclusion}
We propose a comprehensive framework for generating precise height maps at a global scale, utilizing both Sentinel and GEDI data for training. Our framework is built upon fully convolutional neural networks and utilizes cloud-free satellite imagery composites, a novel loss function designed to mitigate GEDI geolocation inaccuracies, and several preprocessing steps.
It enables the generation of high-resolution ($10$~m) global-scale height maps that stand on par with or even surpass the quality of specialized regional height map products. Such height maps play a crucial role in understanding Earth's carbon dynamics and are instrumental in addressing climate change mitigation efforts.

\section*{Acknowledgements}
This work is supported via the AI4Forest project, which is funded by the German Federal Ministry of Education and Research (BMBF; grant number 01IS23025A) and the French National Research Agency~(ANR). We gratefully acknowledge the substantial computational resources that were made available to us by the PALMA II cluster at the University of Münster (subsidized by the DFG; INST 211/667-1) as well as by the Zuse Institute Berlin. We also appreciate the hardware donation of an A100 Tensor Core GPU from Nvidia. Additionally, we extend our gratitude to Jorunn Mense, Karsten Schrödter, and Julian Kranz for their valuable feedback on the final version of the manuscript. This work was further supported by CTrees (\url{https://ctrees.org}).

\newpage

\section*{Impact Statement}
This work aims to enhance the accuracy of forest monitoring through machine learning techniques. Currently, forests absorb approximately half of the CO2 emissions generated by human activities and play a crucial role in mitigating global warming~\citep{friedlingsteinGlobalCarbonBudget2022}.
However, they are increasingly vulnerable to the impacts of climate change and direct land use changes, such as deforestation and degradation~\citep{anderegg2022climate}. 
In line with the UN's Sustainable Development Goals~(SDGs)\footnote{\url{https://sdgs.un.org/2030agenda}}, the Bonn Challenge on forests\footnote{\url{https://www.bonnchallenge.org}}, and the Glasgow Declaration of halting forest loss, 
managing and conserving forests is an indispensable component of climate adaptation and mitigation strategies. Moreover, numerous private entities engage in carbon credit investments, financing projects for forest growth and tree planting within the voluntary carbon market. Accurate monitoring of these projects is essential to ensure their additional value and permanence, thereby mitigating potential leakage effects.\footnote{\url{https://www.theguardian.com/environment/2023/jan/18/revealed-forest-carbon-offsets-biggest-provider-worthless-verra-aoe}}

Forest conservation, regeneration, afforestation, and reforestation are key ingredients of climate mitigation policies in future scenarios that meet the goals of the Paris Agreement on climate.\footnote{\url{https://www.ipcc.ch/report/sixth-assessment-report-working group-3/}} Up to now, the means to assess forest carbon stocks and their changes over time has been through labor-intensive field inventories. Although inventories provide robust estimates of carbon stocks at national scale, their sampling is too sparse to allow monitoring of regional and local changes, and their infrequent revisit cycle does allow to capture shocks causing abrupt forest loss like fires, insects attacks, windthrown events. Moreover, most tropical forested countries do not have an inventory and use default growth rates and simple methods to estimate their forest carbon changes~\citep{grassi2022carbon}.

Precise and up-to-date wall-to-wall high-resolution information about the structure, health, and carbon stocks of forests in the world is critical to assess the current state of forest carbon sinks, predict their future state, and trigger appropriate measures against forest loss. The framework developed in this work aids in providing a consistent and global height map utilizing publicly available satellite data from optical, radar, and GEDI spaceborne sensors. This map can be used as a reference to derive more accurate biomass carbon stocks, as a baseline for 
policymakers to make more accurate decisions about forest management, and further related policies.

\bibliographystyle{icml2024}
\bibliography{library_jan}

\begin{thebibliography}{37}
\providecommand{\natexlab}[1]{#1}
\providecommand{\url}[1]{\texttt{#1}}
\expandafter\ifx\csname urlstyle\endcsname\relax
  \providecommand{\doi}[1]{doi: #1}\else
  \providecommand{\doi}{doi: \begingroup \urlstyle{rm}\Url}\fi

\bibitem[Anderegg et~al.(2022)Anderegg, Wu, Acil, Carvalhais, Pugh, Sadler, and Seidl]{anderegg2022climate}
Anderegg, W.~R., Wu, C., Acil, N., Carvalhais, N., Pugh, T.~A., Sadler, J.~P., and Seidl, R.
\newblock A climate risk analysis of {Earth}’s forests in the 21st century.
\newblock \emph{Science}, 377\penalty0 (6610):\penalty0 1099--1103, 2022.

\bibitem[Braaten(2024)]{justinbraatenSentinel2CloudMasking}
Braaten, J.
\newblock {Sentinel-2 Cloud Masking} with s2cloudless.
\newblock \url{https://developers.google.com/earth-engine/tutorials/community/sentinel-2-s2cloudless}, 2024.
\newblock Accessed 2024-05-15.

\bibitem[Chaurasia \& Culurciello(2017)Chaurasia and Culurciello]{linknet}
Chaurasia, A. and Culurciello, E.
\newblock Linknet: Exploiting encoder representations for efficient semantic segmentation.
\newblock In \emph{2017 IEEE Visual Communications and Image Processing (VCIP)}, pp.\  1--4. IEEE, 2017.

\bibitem[Chen et~al.(2017)Chen, Papandreou, Schroff, and Adam]{deeplabv3}
Chen, L.-C., Papandreou, G., Schroff, F., and Adam, H.
\newblock Rethinking atrous convolution for semantic image segmentation.
\newblock \emph{arXiv preprint arXiv:1706.05587}, 2017.

\bibitem[Chen et~al.(2018)Chen, Zhu, Papandreou, Schroff, and Adam]{deeplabv3plus}
Chen, L.-C., Zhu, Y., Papandreou, G., Schroff, F., and Adam, H.
\newblock Encoder-decoder with atrous separable convolution for semantic image segmentation.
\newblock In \emph{Proceedings of the European Conference on Computer Vision (ECCV)}, pp.\  801--818, 2018.

\bibitem[Deng et~al.(2009)Deng, Dong, Socher, Li, Li, and Fei-Fei]{ImageNet}
Deng, J., Dong, W., Socher, R., Li, L.-J., Li, K., and Fei-Fei, L.
\newblock Imagenet: A large-scale hierarchical image database.
\newblock In \emph{2009 IEEE Conference on Computer Vision and Pattern Recognition}, pp.\  248--255, 2009.

\bibitem[Dubayah et~al.(2020)Dubayah, Blair, Goetz, Fatoyinbo, Hansen, Healey, Hofton, Hurtt, Kellner, Luthcke, et~al.]{dubayahGlobalEcosystemDynamics2020}
Dubayah, R., Blair, J.~B., Goetz, S., Fatoyinbo, L., Hansen, M., Healey, S., Hofton, M., Hurtt, G., Kellner, J., Luthcke, S., et~al.
\newblock The global ecosystem dynamics investigation: High-resolution laser ranging of the {Earth}'s forests and topography.
\newblock \emph{Science of Remote Sensing}, 1:\penalty0 100002, 2020.

\bibitem[{European Space Agency}(2024{\natexlab{a}})]{ESA_Sentinel1}
{European Space Agency}.
\newblock Sentinel-1 -- radar vision for {Copernicus}.
\newblock \url{https://www.esa.int/Applications/Observing_the_Earth/Copernicus/Sentinel-1}, 2024{\natexlab{a}}.
\newblock Accessed: 2024-05-15.

\bibitem[{European Space Agency}(2024{\natexlab{b}})]{ESA_Sentinel2}
{European Space Agency}.
\newblock Sentinel-2 -- colour vision for {Copernicus}.
\newblock \url{https://www.esa.int/Applications/Observing_the_Earth/Copernicus/Sentinel-2}, 2024{\natexlab{b}}.
\newblock Accessed: 2024-05-15.

\bibitem[Fan et~al.(2020)Fan, Wang, Li, and Wang]{manet}
Fan, T., Wang, G., Li, Y., and Wang, H.
\newblock {Ma-Net}: A multi-scale attention network for liver and tumor segmentation.
\newblock \emph{IEEE Access}, 8:\penalty0 179656--179665, 2020.

\bibitem[Fayad et~al.(2023)Fayad, Ciais, Schwartz, Wigneron, Baghdadi, de~Truchis, d'Aspremont, Frappart, Saatchi, Pellissier-Tanon, and Bazzi]{fayadVisionTransformersNew2023}
Fayad, I., Ciais, P., Schwartz, M., Wigneron, J.-P., Baghdadi, N., de~Truchis, A., d'Aspremont, A., Frappart, F., Saatchi, S., Pellissier-Tanon, A., and Bazzi, H.
\newblock Vision transformers, a new approach for high-resolution and large-scale mapping of canopy heights.
\newblock \emph{arXiv preprint arXiv:2304.11487}, 2023.

\bibitem[Fort et~al.(2019)Fort, Hu, and Lakshminarayanan]{Fort2019}
Fort, S., Hu, H., and Lakshminarayanan, B.
\newblock Deep ensembles: A loss landscape perspective.
\newblock \emph{arXiv preprint arXiv:1912.02757}, 2019.

\bibitem[Friedlingstein et~al.(2022)Friedlingstein, Jones, O'Sullivan, Andrew, Bakker, Hauck, Le~Qu\'er\'e, Peters, Peters, Pongratz, Sitch, Canadell, Ciais, Jackson, Alin, Anthoni, Bates, Becker, Bellouin, Bopp, Chau, Chevallier, Chini, Cronin, Currie, Decharme, Djeutchouang, Dou, Evans, Feely, Feng, Gasser, Gilfillan, Gkritzalis, Grassi, Gregor, Gruber, G\"urses, Harris, Houghton, Hurtt, Iida, Ilyina, Luijkx, Jain, Jones, Kato, Kennedy, Klein~Goldewijk, Knauer, Korsbakken, K\"ortzinger, Landsch\"utzer, Lauvset, Lef\`evre, Lienert, Liu, Marland, McGuire, Melton, Munro, Nabel, Nakaoka, Niwa, Ono, Pierrot, Poulter, Rehder, Resplandy, Robertson, R\"odenbeck, Rosan, Schwinger, Schwingshackl, S\'ef\'erian, Sutton, Sweeney, Tanhua, Tans, Tian, Tilbrook, Tubiello, van~der Werf, Vuichard, Wada, Wanninkhof, Watson, Willis, Wiltshire, Yuan, Yue, Yue, Zaehle, and Zeng]{friedlingsteinGlobalCarbonBudget2022}
Friedlingstein, P., Jones, M.~W., O'Sullivan, M., Andrew, R.~M., Bakker, D. C.~E., Hauck, J., Le~Qu\'er\'e, C., Peters, G.~P., Peters, W., Pongratz, J., Sitch, S., Canadell, J.~G., Ciais, P., Jackson, R.~B., Alin, S.~R., Anthoni, P., Bates, N.~R., Becker, M., Bellouin, N., Bopp, L., Chau, T. T.~T., Chevallier, F., Chini, L.~P., Cronin, M., Currie, K.~I., Decharme, B., Djeutchouang, L.~M., Dou, X., Evans, W., Feely, R.~A., Feng, L., Gasser, T., Gilfillan, D., Gkritzalis, T., Grassi, G., Gregor, L., Gruber, N., G\"urses, O., Harris, I., Houghton, R.~A., Hurtt, G.~C., Iida, Y., Ilyina, T., Luijkx, I.~T., Jain, A., Jones, S.~D., Kato, E., Kennedy, D., Klein~Goldewijk, K., Knauer, J., Korsbakken, J.~I., K\"ortzinger, A., Landsch\"utzer, P., Lauvset, S.~K., Lef\`evre, N., Lienert, S., Liu, J., Marland, G., McGuire, P.~C., Melton, J.~R., Munro, D.~R., Nabel, J. E. M.~S., Nakaoka, S.-I., Niwa, Y., Ono, T., Pierrot, D., Poulter, B., Rehder, G., Resplandy, L., Robertson, E., R\"odenbeck, C., Rosan, T.~M., Schwinger,
  J., Schwingshackl, C., S\'ef\'erian, R., Sutton, A.~J., Sweeney, C., Tanhua, T., Tans, P.~P., Tian, H., Tilbrook, B., Tubiello, F., van~der Werf, G.~R., Vuichard, N., Wada, C., Wanninkhof, R., Watson, A.~J., Willis, D., Wiltshire, A.~J., Yuan, W., Yue, C., Yue, X., Zaehle, S., and Zeng, J.
\newblock Global carbon budget 2021.
\newblock \emph{Earth System Science Data}, 14\penalty0 (4):\penalty0 1917--2005, 2022.

\bibitem[Ganaie et~al.(2022)Ganaie, Hu, Malik, Tanveer, and Suganthan]{Ganaie2021}
Ganaie, M., Hu, M., Malik, A., Tanveer, M., and Suganthan, P.
\newblock Ensemble deep learning: A review.
\newblock \emph{Engineering Applications of Artificial Intelligence}, 115:\penalty0 105151, 2022.

\bibitem[Grassi et~al.(2022)Grassi, Conchedda, Federici, Abad Vi\~nas, Korosuo, Melo, Rossi, Sandker, Somogyi, Vizzarri, and Tubiello]{grassi2022carbon}
Grassi, G., Conchedda, G., Federici, S., Abad Vi\~nas, R., Korosuo, A., Melo, J., Rossi, S., Sandker, M., Somogyi, Z., Vizzarri, M., and Tubiello, F.~N.
\newblock Carbon fluxes from land 2000--2020: bringing clarity to countries' reporting.
\newblock \emph{Earth System Science Data}, 14\penalty0 (10):\penalty0 4643--4666, 2022.

\bibitem[He et~al.(2016)He, Zhang, Ren, and Sun]{heDeepResidualLearning2015}
He, K., Zhang, X., Ren, S., and Sun, J.
\newblock Deep residual learning for image recognition.
\newblock In \emph{2016 IEEE Conference on Computer Vision and Pattern Recognition (CVPR)}, pp.\  770--778, 2016.

\bibitem[Hu et~al.(2020)Hu, Zhang, Su, Zheng, Lin, and Guo]{huMappingGlobalMangrove2020}
Hu, T., Zhang, Y., Su, Y., Zheng, Y., Lin, G., and Guo, Q.
\newblock Mapping the global mangrove forest aboveground biomass using multisource remote sensing data.
\newblock \emph{Remote Sensing}, 12\penalty0 (10), 2020.

\bibitem[Izmailov et~al.(2018)Izmailov, Podoprikhin, Garipov, Vetrov, and Wilson]{Izmailov2018}
Izmailov, P., Podoprikhin, D., Garipov, T., Vetrov, D., and Wilson, A.~G.
\newblock Averaging weights leads to wider optima and better generalization.
\newblock \emph{arXiv preprint arXiv:1803.05407}, 2018.

\bibitem[Kacic et~al.(2023)Kacic, Thonfeld, Gessner, and Kuenzer]{kacicForestStructureCharacterization2023}
Kacic, P., Thonfeld, F., Gessner, U., and Kuenzer, C.
\newblock Forest structure characterization in {Germany}: Novel products and analysis based on {GEDI}, {Sentinel-1} and {Sentinel-2} data.
\newblock \emph{Remote Sensing}, 15\penalty0 (8), 2023.

\bibitem[Lang et~al.(2023)Lang, Jetz, Schindler, and Wegner]{langHighresolutionCanopyHeight2023}
Lang, N., Jetz, W., Schindler, K., and Wegner, J.~D.
\newblock A high-resolution canopy height model of the {Earth}.
\newblock \emph{Nature Ecology {\&} Evolution}, 7\penalty0 (11):\penalty0 1778--1789, 2023.

\bibitem[Li et~al.(2018)Li, Xiong, An, and Wang]{pan_net}
Li, H., Xiong, P., An, J., and Wang, L.
\newblock Pyramid attention network for semantic segmentation.
\newblock \emph{arXiv preprint arXiv:1805.10180}, 2018.

\bibitem[Lin et~al.(2017)Lin, Doll{\'a}r, Girshick, He, Hariharan, and Belongie]{fpn_net}
Lin, T.-Y., Doll{\'a}r, P., Girshick, R., He, K., Hariharan, B., and Belongie, S.
\newblock Feature pyramid networks for object detection.
\newblock In \emph{Proceedings of the IEEE Conference on Computer Vision and Pattern Recogfnition}, pp.\  2117--2125, 2017.

\bibitem[Liu et~al.(2023)Liu, Brandt, Nord-Larsen, Chave, Reiner, Lang, Tong, Ciais, Igel, Pascual, Guerra-hernandez, Li, Mugabowindekwe, Saatchi, Yue, Chen, and Fensholt]{liuOverlookedContributionTrees2023}
Liu, S., Brandt, M., Nord-Larsen, T., Chave, J., Reiner, F., Lang, N., Tong, X., Ciais, P., Igel, C., Pascual, A., Guerra-hernandez, J., Li, S., Mugabowindekwe, M., Saatchi, S., Yue, Y., Chen, Z., and Fensholt, R.
\newblock The overlooked contribution of trees outside forests to tree cover and woody biomass across {Europe}.
\newblock \emph{Science Advances}, 9\penalty0 (37), 2023.

\bibitem[Loshchilov \& Hutter(2017)Loshchilov and Hutter]{AdamW}
Loshchilov, I. and Hutter, F.
\newblock Fixing weight decay regularization in {Adam}.
\newblock \emph{CoRR}, abs/1711.05101, 2017.

\bibitem[Potapov et~al.(2021)Potapov, Li, Hernandez-Serna, Tyukavina, Hansen, Kommareddy, Pickens, Turubanova, Tang, Silva, Armston, Dubayah, Blair, and Hofton]{potapovMappingGlobalForest2021}
Potapov, P., Li, X., Hernandez-Serna, A., Tyukavina, A., Hansen, M.~C., Kommareddy, A., Pickens, A., Turubanova, S., Tang, H., Silva, C.~E., Armston, J., Dubayah, R., Blair, J.~B., and Hofton, M.
\newblock Mapping global forest canopy height through integration of {GEDI} and {Landsat} data.
\newblock \emph{Remote Sensing of Environment}, 253:\penalty0 112165, 2021.

\bibitem[Ronneberger et~al.(2015)Ronneberger, Fischer, and Brox]{ronnebergerUNetConvolutionalNetworks2015}
Ronneberger, O., Fischer, P., and Brox, T.
\newblock {U-Net}: Convolutional networks for biomedical image segmentation.
\newblock In \emph{Medical Image Computing and Computer-Assisted Intervention -- MICCAI 2015}, pp.\  234--241, 2015.

\bibitem[Schleich et~al.(2023)Schleich, Durrieu, Soma, and Vega]{schleichImprovingGEDIFootprint2023}
Schleich, A., Durrieu, S., Soma, M., and Vega, C.
\newblock Improving {GEDI} footprint geolocation using a high-resolution digital elevation model.
\newblock \emph{IEEE Journal of Selected Topics in Applied Earth Observations and Remote Sensing}, 16:\penalty0 7718--7732, 2023.

\bibitem[Schwartz et~al.(2023)Schwartz, Ciais, De~Truchis, Chave, Ottlé, Vega, Wigneron, Nicolas, Jouaber, Liu, Brandt, and Fayad]{schwartzFORMSForestMultiple2023}
Schwartz, M., Ciais, P., De~Truchis, A., Chave, J., Ottlé, C., Vega, C., Wigneron, J.-P., Nicolas, M., Jouaber, S., Liu, S., Brandt, M., and Fayad, I.
\newblock {FORMS}: Forest multiple source height, wood volume, and biomass maps in {France} at 10 to 30 m resolution based on {Sentinel-1}, {Sentinel-2}, and {Global Ecosystem Dynamics Investigation} ({GEDI}) data with a deep learning approach.
\newblock \emph{Earth System Science Data}, 15\penalty0 (11):\penalty0 4927--4945, 2023.

\bibitem[Schwartz et~al.(2024)Schwartz, Ciais, Ottl{\'e}, de~Truchis, Vega, Fayad, Brandt, Fensholt, Baghdadi, Morneau, Morin, Guyon, Dayau, and Wigneron]{schwartzHighresolutionCanopyHeight2022}
Schwartz, M., Ciais, P., Ottl{\'e}, C., de~Truchis, A., Vega, C., Fayad, I., Brandt, M., Fensholt, R., Baghdadi, N., Morneau, F., Morin, D., Guyon, D., Dayau, S., and Wigneron, J.-P.
\newblock {High-resolution canopy height map in the {Landes} forest ({France}) based on {GEDI}, {Sentinel-1}, and {Sentinel-2} data with a deep learning approach}.
\newblock \emph{{International Journal of Applied Earth Observation and Geoinformation}}, 128:\penalty0 103711, 2024.

\bibitem[Sloan \& Sayer(2015)Sloan and Sayer]{sloanForestResourcesAssessment2015}
Sloan, S. and Sayer, J.~A.
\newblock Forest resources assessment of 2015 shows positive global trends but forest loss and degradation persist in poor tropical countries.
\newblock \emph{Forest Ecology and Management}, 352:\penalty0 134--145, 2015.

\bibitem[Smith(2017)]{Smith2015}
Smith, L.~N.
\newblock Cyclical learning rates for training neural networks.
\newblock In \emph{2017 IEEE Winter Conference on Applications of Computer Vision (WACV)}, pp.\  464--472. IEEE, 2017.

\bibitem[Tang et~al.(2023)Tang, Stoker, Luthcke, Armston, Lee, Blair, and Hofton]{tangEvaluatingMitigatingImpact2023}
Tang, H., Stoker, J., Luthcke, S., Armston, J., Lee, K., Blair, B., and Hofton, M.
\newblock Evaluating and mitigating the impact of systematic geolocation error on canopy height measurement performance of {GEDI}.
\newblock \emph{Remote Sensing of Environment}, 291:\penalty0 113571, 2023.

\bibitem[Williams et~al.(2006)Williams, Goward, and Arvidson]{Landsat}
Williams, D.~L., Goward, S., and Arvidson, T.
\newblock Landsat.
\newblock \emph{Photogrammetric Engineering \& Remote Sensing}, 72\penalty0 (10):\penalty0 1171--1178, 2006.

\bibitem[Zhao et~al.(2017)Zhao, Shi, Qi, Wang, and Jia]{pspnet}
Zhao, H., Shi, J., Qi, X., Wang, X., and Jia, J.
\newblock Pyramid scene parsing network.
\newblock In \emph{Proceedings of the IEEE Conference on Computer Vision and Pattern Recognition}, pp.\  2881--2890, 2017.

\bibitem[Zhou et~al.(2019)Zhou, Siddiquee, Tajbakhsh, and Liang]{unetplusplus}
Zhou, Z., Siddiquee, M. M.~R., Tajbakhsh, N., and Liang, J.
\newblock {UNet++}: Redesigning skip connections to exploit multiscale features in image segmentation.
\newblock \emph{IEEE Transactions on Medical Imaging}, 39\penalty0 (6):\penalty0 1856--1867, 2019.

\bibitem[Zimmer et~al.(2023)Zimmer, Spiegel, and Pokutta]{Zimmer2021}
Zimmer, M., Spiegel, C., and Pokutta, S.
\newblock {H}ow {I} {L}earned {T}o {S}top {W}orrying {A}nd {L}ove {R}etraining.
\newblock In \emph{International Conference on Learning Representations}, 2023.

\bibitem[Zimmer et~al.(2024)Zimmer, Spiegel, and Pokutta]{Zimmer2023}
Zimmer, M., Spiegel, C., and Pokutta, S.
\newblock Sparse model soups: A recipe for improved pruning via model averaging.
\newblock In \emph{International Conference on Learning Representations}, 2024.

\end{thebibliography}

% %%%%%%%%%%%%%%%%%%%%%%%%%%%%%%%%%%%%%%%%%%%%%%%%%%%%%%%%%%%%%%%%%%%%%%%%%%%%%%%
% %%%%%%%%%%%%%%%%%%%%%%%%%%%%%%%%%%%%%%%%%%%%%%%%%%%%%%%%%%%%%%%%%%%%%%%%%%%%%%%
% % APPENDIX
% %%%%%%%%%%%%%%%%%%%%%%%%%%%%%%%%%%%%%%%%%%%%%%%%%%%%%%%%%%%%%%%%%%%%%%%%%%%%%%%
% %%%%%%%%%%%%%%%%%%%%%%%%%%%%%%%%%%%%%%%%%%%%%%%%%%%%%%%%%%%%%%%%%%%%%%%%%%%%%%%
\newpage
\appendix
\onecolumn

\section{Data}
\subsection{Sentinel-1}
For Sentinel-1, we use all available images from the Sentinel-1 Ground Range Detected~(GRD) dataset, taken between April and October 2020 for the northern hemisphere, and between October 2019 and April 2020 for the southern hemisphere~\cite{ESA_Sentinel1}. This difference in time frame accounts for the varying seasonality. Summer leaf-on images are particularly useful for estimating canopy height, so we aim to obtain a summer leaf-on composite for each hemisphere.

We use Sentinel-1 satellite images acquired in both VV (Vertical Transmit, Vertical Receive) and VH (Vertical Transmit, Horizontal Receive) polarization. Given the substantial differences in the data acquired during ascending and descending satellite orbits, we categorically separate the images from different orbits.
Although Sentinel-1 is mostly unaffected by clouds and general weather, using single images as input data for the model is insufficient. Sentinel-1 images often encounter issues with noise (e.g., speckle or thermal noise), and heavy rain can disrupt image quality. We follow a common approach for addressing these issues (which we follow) is to use the temporal per-pixel median, which effectively eliminates most of the noise problems. By following this approach, we generate four distinct channels: VV from ascending orbits, VV from descending orbits, VH from ascending orbits, and VH from descending orbits. 

\subsection{Sentinel-2}
\label{cloudmasking}
For Sentinel-2, there are different data products available. We use all Sentinel-2 bottom-of-the-atmosphere~(BOA) images from the same time frame as the Sentinel-1 data~\cite{ESA_Sentinel2}. Typically, one resorts to a similar temporal composition approach for Sentinel-2 as for Sentinel-1 data. However, such temporal per-pixel median composites are often insufficient for Sentinel-2 data, since the corresponding optical sensor cannot penetrate clouds. Hence, while taking the temporal per-pixel median might yield cloud-free image composites for most regions, such an approach is generally not suited for rainforest regions, where the cloud coverage is often persistent throughout the year. These regions are crucial though for carbon monitoring as they store a significant portion of global carbon. Therefore, we deviate from the common procedure and make use of the following steps to generate Sentinel-2 image composites:
\begin{enumerate}
    \item Instead of directly taking the temporal per-pixel median, we first filter out clouds in each image before aggregating the remaining pixels. We do this by adapting an approach from \citet{justinbraatenSentinel2CloudMasking}. More precisely, each Sentinel-2 image comes with a so-called cloud mask that contains the (estimated) cloud cover/probability for each pixel in that image. We make use of this mask to filter out images with less than 10\% cloud-free pixels. %, as we cannot obtain any useful information from images with 100\% cloud coverage.
    \item We then employ multiple steps to identify and remove disturbed pixels. First, we identify and remove pixels with a cloud probability larger than 30\%. 
    Knowing the sun angle at the time of image capture, we examine each cloud pixel up to 1 km in the (opposite) direction of the sun and search for potential shadow/dark pixels. 
    To identify those dark pixels, we make use of the near-infrared band B8 and the scene-classification band SCL.
    We then remove, for each pixel identified as cloud or shadow, all neighbored pixels within a $300$~m radius.
    \item Finally, we aggregate all the remaining pixels for each image using a temporal per-pixel median.   
\end{enumerate}
The approach described above generally yields reasonable image composites containing no ``gaps''. For very few areas, no valid pixels might remain, thus leading to ``empty'' regions in the input data, see the last row of Figure~\ref{fig:detailed_examples}.

\subsection{GEDI}
To enhance the quality of the GEDI measurements, we apply several filters:
\begin{enumerate}
    \item \textit{$beam\_int > 5$}: The GEDI instrument captures data via $8$ tracks, $4$ of those are measured by only one beam, the so-called ``coverage'' beam. This beam has lower power, making its measurements more susceptible to noise.
    For this reason, we discard those tracks and only resort to the measurements of the remaining so-called ``power'' beams.
    \item \textit{$quality\_flag = 1$}: This flag filters out the GEDI measurements with a poor quality, i.e., those measurements that were significantly disturbed by clouds or adverse weather conditions.
    \item \textit{$solar\_elevation < 0$}: To minimize the impact of solar radiation on GEDI measurements, we exclude all daytime measurements, retaining only those taken during nighttime.
\end{enumerate}

\subsection{SRTM}
\label{appendix:srtm}

GEDI measurements might be incorrect for mountainous regions (cf. \cref{fig:gedi_slope}). GEDI determines (vegetation) height by calculating the time difference between the first and the last of the returned signals. For example, by multiplying a time difference of $112$~ns between the first and the last signal with the speed of light ($15$ cm/ns) results in a height of about $16.8$~m. On slopes, the first and last signals may not originate from the same object/tree, but from nearby objects/trees of \emph{different} heights, leading to an overestimation of the canopy height. Additionally, even in the absence of any object/vegetation, the first and last signals may be reflected at different heights, causing GEDI to register a positive height for these areas as well.

Ideally, the exact slope at each measurement location should be determined using a high-resolution digital terrain model~(DTM), which would allow to identify and exclude these faulty measurements. However, such DTMs are not publicly available at a global scale. Therefore, we resort to data of the SRTM mission, which mapped the Earth's surface height with a resolution of about $30$~m. Although surface height data are not perfect in this context (since forest borders can also appear as slopes), they currently depict the best publicly available data source for the identification of the critical areas.
To improve the label quality, we filter out any measurement with a SRTM slope of $20$\degree or larger. We calculate the slope by taking the height difference within an area of size $5 \times 5$ pixels (corresponding to $150 \times 150$~m). To ensure no valid GEDI measurement is discarded, we set the filter threshold above the possible slope at forest borders. More precisely, if a forest border has trees of height $40$~m, the slope is approximately $15$\degree. Hence, we set the threshold to $20$\degree, allowing for a height change (excluding the forest border) of up to $15$~m.

\section{Training/Validation/Test}
\label{appendix:train_val_test}
\begin{figure}[t]
  \centering
  \vskip 0.15in
  \mycolorbox{\resizebox{0.98\textwidth}{!}{\includegraphics{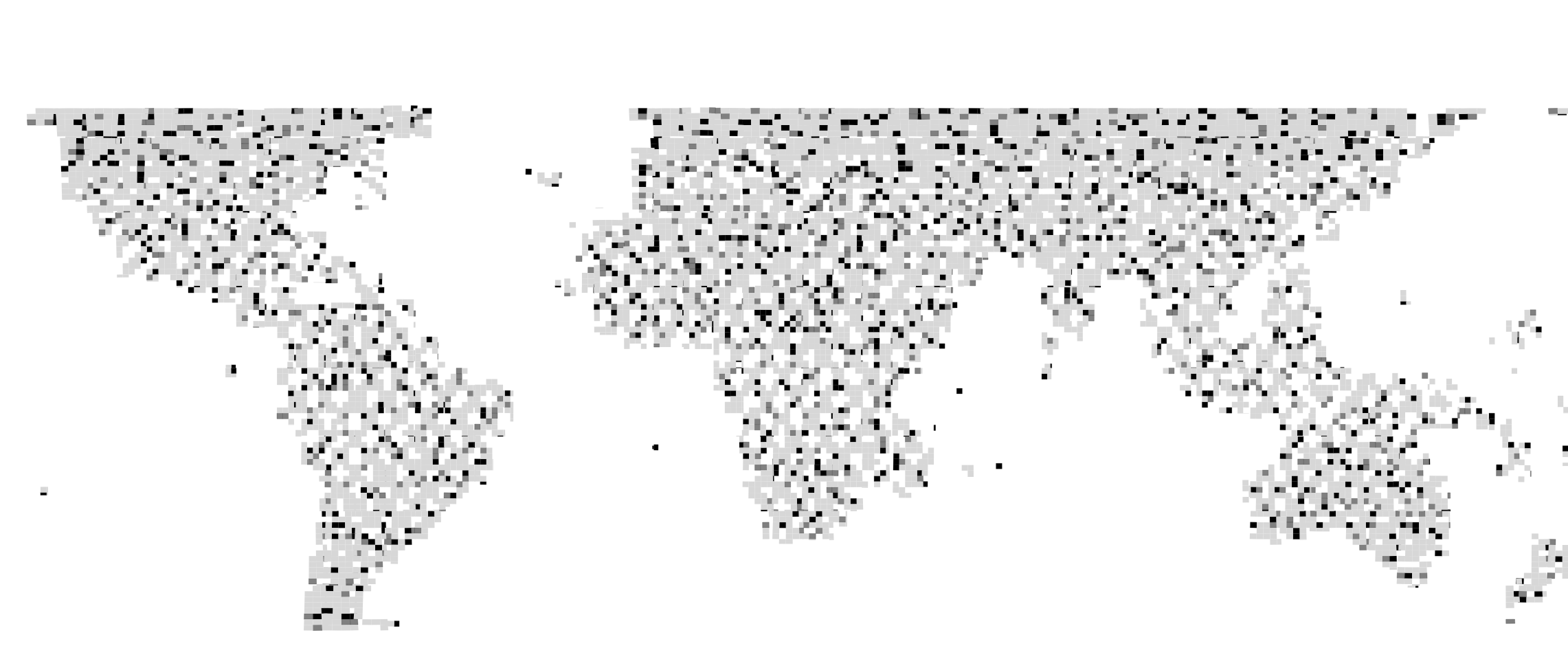}}}
  \caption{Training (light grey), validation (dark grey), and test (black) regions.}
  \label{fig:train_split}
\vskip -0.3in
\end{figure}

We split the data into geographically non-overlapping regions for training, validation, and testing. An illustration of the distribution of training, validation, and test instances can be seen in \cref{fig:train_split}. Each Universal Transverse Mercator~(UTM) zone has, relative to its size, the same proportion of training, validation, and test instances. 
Note that the northern regions are not covered due to the ISS orbit not covering this area. 

\section{Model}
\begin{table}[t]
\centering
\caption{Model hyperparameters for fine-tuning.
\label{tab:hyperparameter}
}
\vskip 0.1in

\begin{tabularx}{\linewidth}{l>{\hsize=0.2\hsize}SX
                                >{\hsize=0.8\hsize\arraybackslash}SX}
\toprule

 \textsc{Hyperparameter}  &  \hfil\textsc{Values}  &   \textsc{Description}  \\

\midrule
weight decay & $0, 0.01, 0.001$ & Weight decay of the AdamW optimizer\\ 
shifted loss & $\mathcal{L}_\textit{S}$, $\mathcal{L}_\textit{NS}$ & We consider both the standard non-shifted loss function $\loss_\textit{NS}$ defined in Equation~(\ref{eq:non_shifted_loss}) as well as the shifted variant $\loss_\textit{S}$ defined in Equation~(\ref{eq:shifted_loss}). \\ %\midrule
pixel-wise loss &  $L_1$, $L_2$, Huber & Both $\loss_\textit{NS}$ and $\loss_\textit{S}$ are based on a pixel-wise loss $\loss$. Here, we consider the $L_1$ (mean absolute error), $L_2$ (mean squared error), and the Huber loss as loss functions. For the Huber loss, we set the cutoff parameter $\delta$ to $3$ (meters). \\ %\midrule
model weights & None, ImageNet & We use the ResNet-50~\citep{heDeepResidualLearning2015} backbone and train the weights either from scratch (None) or resort to pre-trained weights (that stem from the ImageNet~\cite{ImageNet} dataset). \\ \bottomrule
\end{tabularx}
\vskip -0.2in
\end{table}
To fine-tune our model, we conduct a comprehensive hyperparameter search, see \cref{tab:hyperparameter}. 
The final model was trained from scratch using the AdamW optimizer with a weight decay of $0.001$, a batch size of $32$, and an initial learning rate of $0.001$. We implemented a linear learning rate warm-up for the first $10$\% of the total iterations, followed by a linear learning rate scheduler for the remaining $90$\%. Additionally, gradient clipping was applied to prevent gradient explosion. Finally, the shifted Huber loss was used during training, i.e., $\loss_\textit{S}$ with the Huber loss as pixel-wise loss function $\loss$ in Equation~(\ref{eq:shifted_loss}).\footnote{Using sample weights for the training instances led to a worse performance. Hence, no sample weights have been used for training.}

\section{Examples}
\label{examples_of_images}
\begin{figure}
    \centering
    \includegraphics[width=0.93\textwidth]{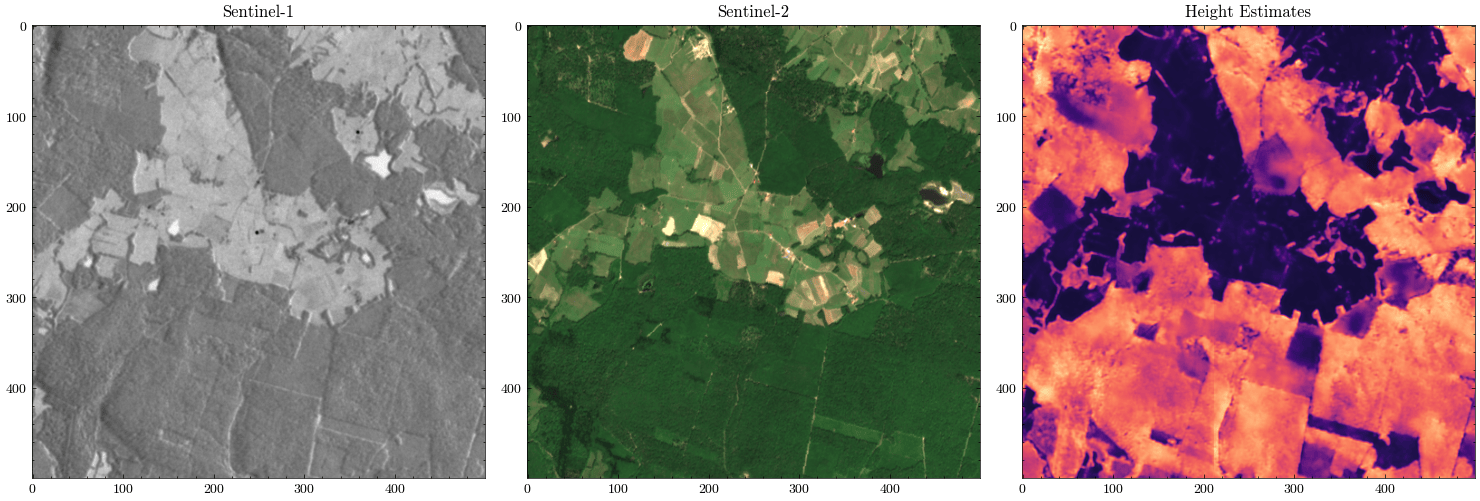}
    \includegraphics[width=0.93\textwidth]{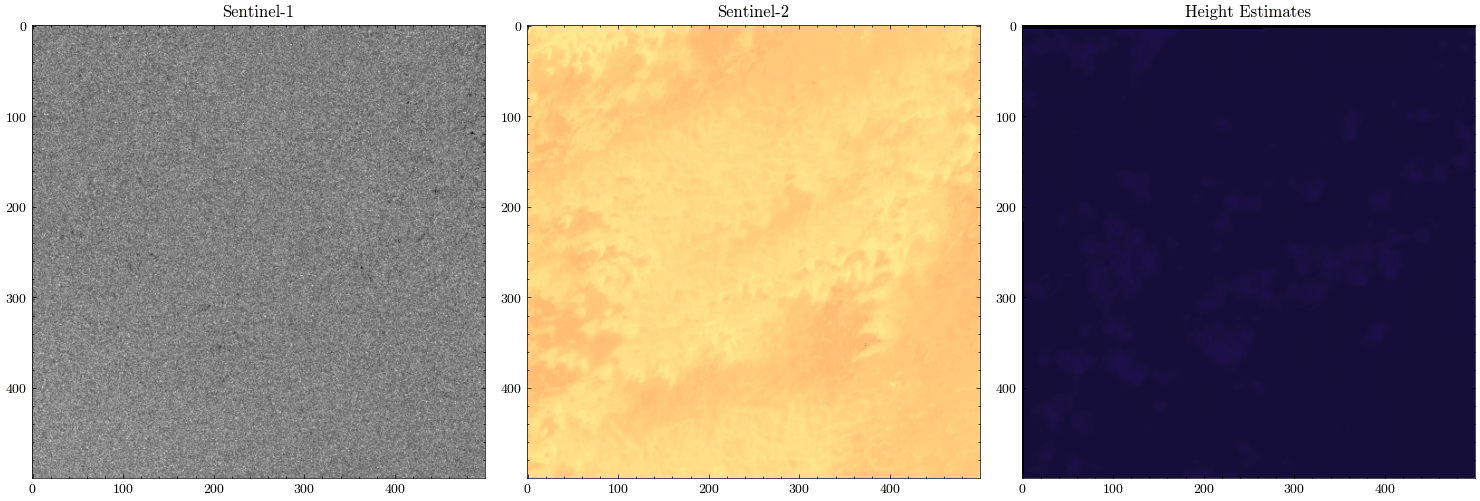}
    \includegraphics[width=0.93\textwidth]{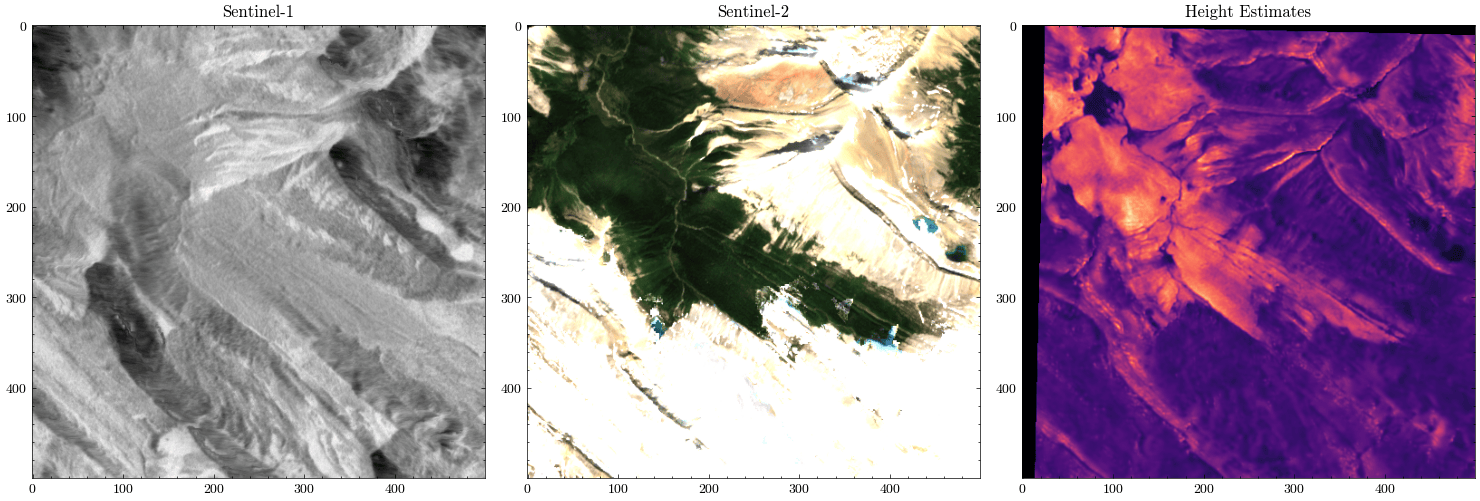}
    \includegraphics[width=0.93\textwidth]{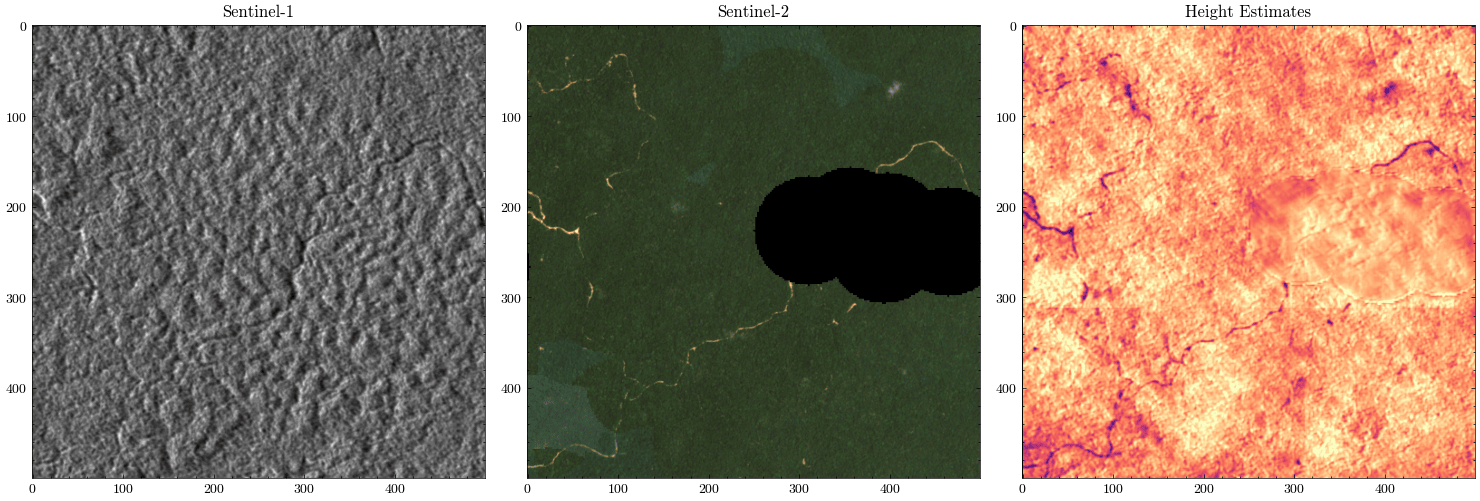}
    \caption{Sentinel-1 and Sentinel-2 images (RGB channels) and our prediction in France (first row), in the Sahara (second row), in Canada (third row), and in the Congo Basin (fourth row; heights from 0 m to 35 m; see \cref{fig:global_map} for the colormap)}
    \label{fig:detailed_examples}
\end{figure}

\cref{fig:detailed_examples} shows Sentinel input data as well as the height estimates induced by our model for some more examples. Note that the last row sketches the (rare) situation in case no input pixels are considered to be valid (see above), leading to an empty area in the composite image. Here, some parts of the Sentinel-2 image were masked out due to a lack of non-cloud pixels. Yet, our model estimates the height reasonably well using the spatial context given in the Sentinel-2 composite as well as the Sentinel-1 data.

\section{Ablation Studies}
We conduct several ablation studies to evaluate the impact of the involved model components.
\subsection{Loss Function}
First, we start by assessing effectiveness of the shifted loss function.
Our objective is to ascertain whether the use of $ \loss_{\textit{S}}$ (see Equation~(\ref{eq:shifted_loss})) enhances the model performance and to identify the most suitable pixel-wise loss function. We compare three commonly used loss functions ($L_1$, $L2$, Huber loss) and conduct experiments using both $\mathcal{L}_{\textit{S}}$ and $ \mathcal{L}_{\textit{NS}}$. 
The results are given in Table~\ref{tab:ablation_loss}. 
All models are trained with the same set of hyperparameters, except for the choice of loss function used during training (first column). We present the results for all metrics (remaining columns) based on the validation set.
One can see that using $\mathcal{L}_{\textit{S}}$ generally leads to equal or better results compared to $\mathcal{L}_{\textit{NS}}$, which was expected since $\mathcal{L}_{\textit{NS}}$ is a special case of $\mathcal{L}_{\textit{S}}$ with $\delta_x = 0$ and $\delta_y = 0$. 

\begin{table}[t]

\centering
\caption{Ablation study on different loss function: $L_1$, $L_2$ and Huber loss both with and without $ \mathcal{L}_{\textit{S}}$. Each row represents the loss function used for training the model, while each column represents the loss function used for evaluating the model's performance. 
}
\label{tab:ablation_loss}
\vskip 0.1in
%\resizebox{\textwidth}{!}{
\begin{tabular}{lcccccc}
\toprule
\textsc{Validation Loss} & \textsc{$L_1$} & \textsc{$L_2$} & {Huber} & \textsc{$ \mathcal{L}_{\textit{S}} + L_1$} & {$ \mathcal{L}_{\textit{S}} + L_2$} & {$ \mathcal{L}_{\textit{S}} + \text{Huber}$} \\
\textsc{Training Loss} \\
\midrule
$L_1$ & 1.68 $\pm$ 0.03 & 11.19 $\pm$ 0.25 &1.04 $\pm$ 0.02 & 1.65 $\pm$ 0.03 & 10.82 $\pm$ 0.27 & 1.01 $\pm$ 0.02 \\
$L_2$ & 1.76 $\pm$ 0.01 & 11.12 $\pm$ 0.12 & 1.05 $\pm$ 0.00 & 1.73 $\pm$ 0.01 & 10.76 $\pm$ 0.13 & 1.02 $\pm$ 0.00 \\
Huber & 1.73 $\pm$ 0.01 & 11.11 $\pm$ 0.18 & 1.04 $\pm$ 0.01 & 1.70 $\pm$ 0.01 & 10.73 $\pm$ 0.20 & 1.01 $\pm$ 0.01\\\midrule
$ \mathcal{L}_{\textit{S}} + L_1$ & 1.68 $\pm$ 0.03 & 11.15 $\pm$ 0.21 & 1.04 $\pm$ 0.02 & 1.64 $\pm$ 0.03 & 10.63 $\pm$ 0.19 & 1.00 $\pm$ 0.02\\
$ \mathcal{L}_{\textit{S}} + L_2$ & 1.75 $\pm$ 0.02 & 11.07 $\pm$ 0.08 & 1.05 $\pm$ 0.01 & 1.71 $\pm$ 0.02 & 10.57 $\pm$ 0.08 & 1.01 $\pm$ 0.01 \\
$ \mathcal{L}_{\textit{S}} + \text{Huber}$ & 1.70 $\pm$ 0.01 & 10.92 $\pm$ 0.16 & 1.03 $\pm$ 0.01 & 1.66 $\pm$ 0.01 & 10.42 $\pm$ 0.17 & 0.99 $\pm$ 0.01 \\
\bottomrule
\end{tabular}
%}
\vskip -0.15in
\end{table}

\subsection{SRTM-Filtering}

\begin{figure}
     \centering
     \includegraphics[width=\textwidth]{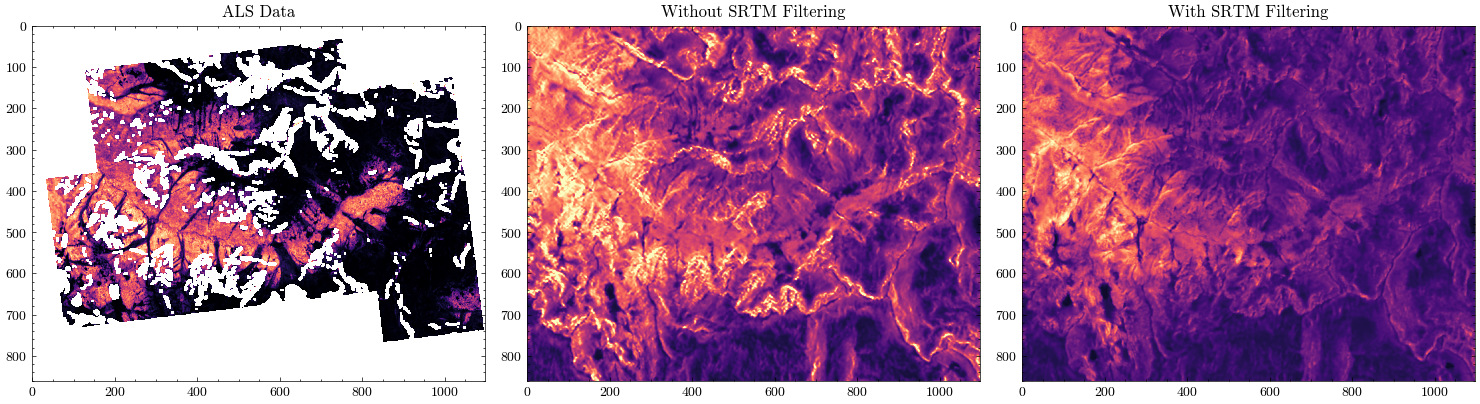}
     \caption{ALS data and both predictions for the ablation area.}     
     \label{fig:ablation_srtm}
\end{figure}

Next, we assess the effectiveness of the SRTM filtering approach. For this, we resort to Airbone Laser Scanning~(ALS) data from the Needle Mountains in the USA (covering an area of about $51.83$~$\text{km}^2$). The ALS data depict (more) precise height measurements and can be considered as ground-truth for this ablation study. We train two models using the same data and the same parameters, except for SRTM filtering.
The MAE is $9.77$~m without SRTM filtering and improves to $7.33$~m with SRTM filtering. An illustration of the ALS data as well as both results is provided in \cref{fig:ablation_srtm}.

\subsection{Architectures}
\begin{table}[t]
\vskip -0.1in
\caption{Ablation study comparing different model architectures showing the shifted Huber loss.}
\label{architecture_validation_loss2}
\centering
\vskip 0.1in
\resizebox{\textwidth}{!}{%
\begin{tabular}{lccccccccc}
\toprule
\textsc{Architecture} & \textsc{DeepLabV3} & \textsc{DeepLabV3+} & \textsc{FPN} & \textsc{LinkNet} &  \textsc{MANet} & \textsc{PAN} & \textsc{PSPNet} & \textsc{UNet} & \textsc{UNet++}\\
\midrule
\textsc{Pre-trained} & $1.22$  & $1.10$  & $1.14$  & $1.06$ & $1.04$  & $1.13$  & $1.13$ & $1.10$ & $1.10$  \\
\textsc{Not Pre-trained} & $1.15$  & $1.08$  & $1.08$ & $1.04$ & $1.03$ & $1.07$  & $1.12$  & $1.01$ & $1.04$ \\
\bottomrule
\end{tabular}
}
\vskip -0.1in
\end{table}
To assess the effectiveness of various deep learning architectures for canopy height prediction, we conducted comparative analyses on several models, including DeepLabV3 \citep{deeplabv3} and V3+ \citep{deeplabv3plus}, FPN \citep{fpn_net}, LinkNet \citep{linknet}, MANet \citep{manet}, PAN \citep{pan_net}, PSPNet \citep{pspnet}, and UNet++ \citep{unetplusplus}. These models were evaluated in both configurations: with and without pre-trained backbones based on ImageNet \citep{ImageNet}.
The comparative analysis, provided in Table~\ref{architecture_validation_loss2}, suggests a strong disparity between different model architectures. Specifically for canopy height prediction, the standard U-Net model~\citep{ronnebergerUNetConvolutionalNetworks2015}, despite its simplicity, outperformed more complex architectures like DeepLabV3~\citep{deeplabv3} or PSPNet~\citep{pspnet}. This outcome underscores the unique demands of canopy height prediction tasks, which may not align with the strengths of architectures designed for broader or different types of image segmentation tasks.
Moreover, the experiment revealed a counter-intuitive finding regarding the use of pre-trained backbones. Contrary to expectations, employing a pre-trained backbone resulted in a decrease in performance across all examined architectures. This decrease was consistent but varied in magnitude among the different models, indicating a potential mismatch between the generalized features learned from ImageNet~\citep{ImageNet} and the specific features of satellite images relevant to canopy height estimation.

\subsection{Learning Rate Scheduler}
\begin{wraptable}{r}{12cm}
\vspace{-20pt}
\caption{Learning rate schedulers with and without amplitude decay (AD).}
\label{tab:cycle_mode_loss}
\centering
%\resizebox{11cm}{!}{%
\vskip 0.1in
\begin{sc}
\begin{tabular}{lccccc}
\toprule
{Mode} & {linear} & {1 cycle} & {2 cycles} & {3 cycles} & {4 cycles} \\
\midrule
Without AD & $-$ & $1.022$ & $1.012$ & $1.010$ & $1.004$ \\
With AD & $1.014$ & $1.016$ & $1.060$ & $1.026$ & $1.053$ \\
\bottomrule
\end{tabular}
\end{sc}
%}
%\vskip -0.1in
\end{wraptable} 

Throughout our experiments, we relied on a linear learning rate scheduler with tuned initial value. The use of cyclical scheduler has previously been found to aid generalization~\citep{Smith2015,Zimmer2021}. To investigate the impact in our setting, we compare two different cyclical schedulers (with and without overall decay of the cycle amplitude) in \Cref{tab:cycle_mode_loss} for different numbers of cycles, using 0.001 as the learning rate bound. These two modes correspond to \texttt{triangular} and \texttt{triangular2} in PyTorch's \texttt{CyclicLR} implementation. We report the shifted Huber loss on the validation set.

\subsection{Ensembles}

\begin{wraptable}{r}{9.5cm}
\vspace{-35pt}
\caption{Single versus ensemble model.}
\label{tab:ablation_ensemble_comparison}
\centering
%\resizebox{\textwidth}{!}{%
\vskip 0.1in
\begin{sc}
\begin{tabular}{lccccc}
\toprule
Model & MAE & MSE & RMSE & MAPE & RRMSE\\
\midrule
Single      & $4.03$ & $36.57$ & $6.05$ & $0.29$ & $0.50$ \\
Ensemble  & $3.94$ & $35.07$ & $5.92$ & $0.28$ & $0.49$ \\
\bottomrule
\end{tabular}
\end{sc}
%}
\vskip -0.1in
\end{wraptable} 
Previous work has shown that the performance of models can be improved by building {ensembles}~\cite{Fort2019, Ganaie2021} or to {parameter averages}~\cite{Izmailov2018, Zimmer2023}) from multiple models. While parameter averages do not increase the inference complexity, more effort is required to find models suitable for averaging (hence, parameter averages are not suited for our case). In Table~\ref{tab:ablation_ensemble_comparison}, we evaluate the single model against an {ensemble} of six models. Here, the ensemble aggregates the estimates of multiple models and, hence, increases the performance in all metrics, albeit at the price of having to evaluate multiple models for inference.

\subsection{Backbone}
\begin{wraptable}{r}{9.5cm}
\vspace{-45pt}
\caption{Different backbones.}
\label{tab:ablation_backbone}
\centering
\vskip 0.1in
\begin{sc}
\begin{tabular}{lccc}
\toprule
 & {ResNet18} & {ResNet50} & {ResNet101}  \\
\midrule
 {Parameters} & \numprint{14362705} & \numprint{32555601} & \numprint{51547729} \\
 {Pre-trained} & $1.117$ & $1.086$ & $1.095$ \\
 {Not Pre-trained} & $1.051$ & $1.017$ & $1.020$ \\
\bottomrule
\end{tabular}
\end{sc}
\vskip -0.2in
\end{wraptable} 
In order to validate our backbone choice, we do an ablation study testing different versions of our backbone and compare them with each other regarding the number of parameters and the shifted Huber loss (on the validation set) for both pre-trained and not pre-trained backbones. The results can be seen in \cref{tab:ablation_backbone}. As already mentioned above, training the backbones from scratch seems to be beneficial in this context, and the ResNet50 backbone seems to be a reasonable choice.

\section{Canopy Height Map}
\label{sec:global_canopy_height_map}
\begin{figure}[t]
  \centering
  \mycolorbox{\resizebox{0.98\textwidth}{!}{\includegraphics{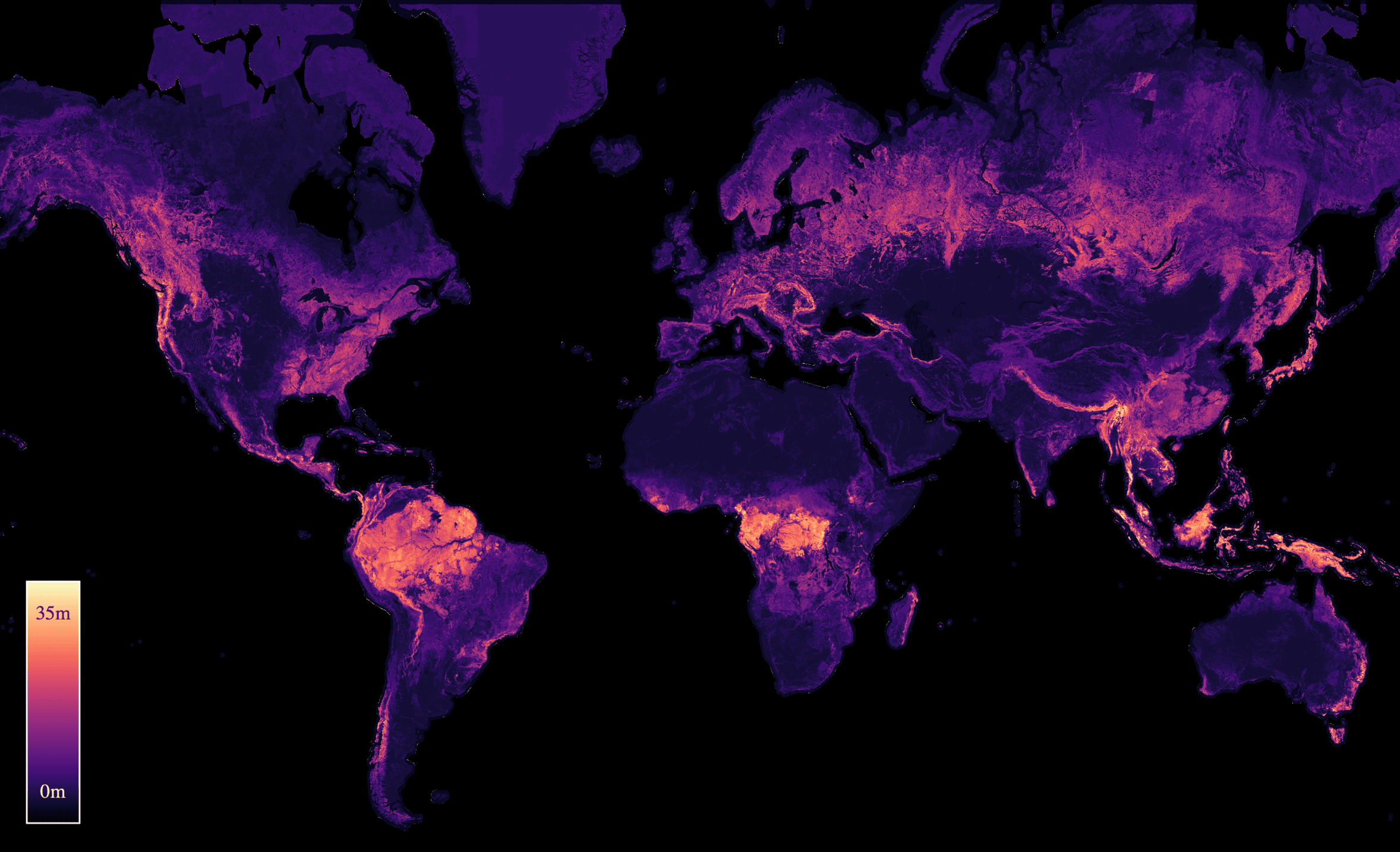}}}
  \vskip -0.1in
  \caption{Global canopy height map (enlarged version of Figure~\ref{fig:global_map}).}
  \label{fig:global_map_large}  
  \vskip -0.1in
\end{figure}

The final canopy height map is shown in Figure~\ref{fig:global_map_large}. For an interactive map, we refer the reader to the  {Google Earth Engine}, which allows to inspect any location on the map in more detail.\footnote{\href{https://worldwidemap.projects.earthengine.app/view/canopy-height-2020}{https://worldwidemap.projects.earthengine.app/view/canopy-height-2020}}
\end{document}